\documentclass[12pt]{article}

\usepackage{graphicx,amsmath,amssymb,url}

\usepackage{calc}
\usepackage{amssymb}
\usepackage{amstext}
\usepackage{amsmath}
\usepackage{amsthm}
\usepackage{apalike}
\usepackage{comment}
\usepackage{wasysym}
\usepackage{subfigure}
\usepackage{graphicx}
\usepackage{algorithmic}
\usepackage{color}
\usepackage{stmaryrd}

\newcommand{\with}{\,  | \,}
\newcommand{\given}{\, | \,}

\newcommand{\set}[1]{\mathcal{#1}}

\newcommand{\algo}{\mathbf{ALG}}
\newcommand{\model}{M}
\newcommand{\modelclass}{\mathbf{M}}
\renewcommand{\vec}[1]{\boldsymbol{#1}}
\newcommand{\Prob}{\mathbf{P}}

\begin{document}


\title{\Large\bf Learning from Imprecise and Fuzzy Observations: 
Data Disambiguation through Generalized\\ Loss Minimization}

\author{Eyke H\"ullermeier\\
Department of Mathematics and Computer Science \\
University of Marburg, Germany \\
eyke@mathematik.uni-marburg.de}


\maketitle

\begin{abstract}
Methods for analyzing or learning from ``fuzzy data'' have attracted increasing attention in recent years. In many cases, however, existing methods (for precise, non-fuzzy data) are extended to the fuzzy case in an ad-hoc manner, and without carefully considering the interpretation of a fuzzy set when being used for modeling data. Distinguishing between an \emph{ontic} and an \emph{epistemic} interpretation of fuzzy set-valued data, and focusing on the latter, we argue that a ``fuzzification'' of learning algorithms based on an application of the generic extension principle is not appropriate. In fact, the extension principle fails to properly exploit the inductive bias underlying statistical and machine learning methods, although this bias, at least in principle, offers a means for ``disambiguating'' the fuzzy data. Alternatively, we therefore propose a method which is based on the generalization of loss functions in empirical risk minimization, and which performs model identification and data disambiguation simultaneously. Elaborating on the fuzzification of specific types of losses, we establish connections to well-known loss functions in regression and classification. We compare our approach with related methods and illustrate its use in logistic regression for binary classification.\\[3mm]

\end{abstract}

\section{Introduction}
\label{Intro}

The learning of models from imprecise data, such as interval data or, more generally, data modeled in terms of fuzzy subsets of an underlying reference space, has gained increasing interest in recent years \cite{sanc_at07,deno_ml11,deno_ml11b,cour_lf11,vier_sm}. Indeed, while problems such as fuzzy regression analysis \cite{diam_fl88,diam_fr98,tana_pd,chan_fr01,gonz_eo09,ferr_al10} have already been studied for a long time, the scope is currently broadening, both in terms of the problems tackled (e.g., classification, clustering, ranking) and the uncertainty formalisms used (e.g., probability distributions, histograms, intervals, fuzzy sets, belief functions).

Needless to say, learning from imprecise and uncertain data also requires the extension of corresponding learning algorithms. Unfortunately, this is often done without clarifying the actual meaning of an uncertain observation, although representations such as intervals or fuzzy sets can obviously be interpreted in different ways. In particular, an \emph{ontic} interpretation of (fuzzy) set-valued data should be carefully distinguished from an \emph{epistemic} one \cite{dubo_ov11}. This difference is reflected, for example, in different approaches to fuzzy statistics, where fuzzy random variables can be formalized in an epistemic \cite{kwaa_fr78,kwaa_fr79,krus_sw}   as well as an ontic way \cite{puri_fr86}; see \cite{cous_ot09} for a comparison of these views in this context. Surprisingly, however, the fact that these two interpretations also call for very different types of extensions of existing learning algorithms and methods for data analysis seems to be largely ignored in the literature. 

Under the ontic view, a variable can assume a fuzzy set as its ``true value''; for example, one may argue that assigning a precise value to the variable ``daily sunshine duration'' is not very meaningful, and that a specification of sunshine durations in terms of intervals or fuzzy sets is more appropriate. This interpretation suggests the learning of models that produce fuzzy sets as predictions, that is to say, models that \emph{reproduce} the observed data. As opposed to this, a reproduction of the data appears less reasonable under the epistemic view, where fuzzy sets are used to describe, not the data itself, but the uncertain or imprecise \emph{knowledge} about the data: A fuzzy set defines a possibility distribution that specifies a degree of plausibility for each potential precise value. As we shall explain in more detail later on, one should then rather try to ``disambiguate'' the data instead of reproducing it. 

The possibilistic interpretation of fuzzy sets in the epistemic case, that we focus on in this paper, naturally suggests a ``fuzzification'' of learning algorithms based on an application of the generic extension principle \cite{cern_ot11,xian_tf13}. As we shall argue, however, this approach is not appropriate and prone to fail in the context of data analysis. The main reason, to be detailed in Section~3, is a lack of differentiation between the possible data instantiations (i.e., the instantiation of each imprecise observation by a precise value). Such a differentiation, however, is typically suggested by the model assumptions through which the learning algorithm justifies its generalization beyond the data observed.

This idea of differentiating between instantiations of the data leads us to the notion of ``data disambiguation'' that we already mentioned above: \emph{When learning from imprecise data under the epistemic view, model identification and data disambiguation should go hand in hand}. To this end, we propose an approach based on the generalization of loss functions in empirical risk minimization.

The rest of the paper is organized as follows. In the next section, we introduce the basic setting that we consider and the main notation that we shall use throughout the paper (see Table \ref{tab:notation} for a summary). In Section~3, we explain the aforementioned problems caused by the use of the extension principle and elaborate on our idea of data disambiguation. Our new approach to learning from fuzzy data based on generalized loss functions is then introduced in Section~4. Section~5 is devoted to a comparison with an alternative and closely related method that was recently introduced by Denoeux \cite{deno_ml11,deno_ml11b}. In Section~6, we illustrate our approach on a concrete learning problem. Finally, we conclude with a summary and some additional remarks in Section~7.  




\section{Notation and Basic Setting}

We consider the problem of \emph{model induction}, which, roughly speaking, consists of passing from a specific data sample to a general (though hypothetical) model describing the data generating process or at least certain properties of this process. In this setting, a learning (data analysis) algorithm $\algo$ is given as input a set 
\begin{equation}\label{eq:precisedata}
\set{D} = \big\{ \,  \vec{z}_i  \, \big\}_{i=1}^N \, \in \, \set{Z}^N
\end{equation}
of data points $\vec{z}_i  \in \set{Z}$. As output, the algorithm produces a model $\model \in \modelclass$, where $\modelclass$ is a predefined model class. Formally, the algorithm can hence be seen as a mapping
\begin{equation}\label{eq:amap}
\algo: \, \mathbf{D} \rightarrow \modelclass \enspace ,
\end{equation}
where $\mathbf{D}$ is the space of potentially observable data samples. 
For instance, the data points might be vectors in $\set{Z} = \mathbb{R}^d$, and the model could be a partitioning of the data into a finite set of disjoint groups (clusters). Or, the model could be a probability density function characterizing the underlying data generating process. In fact, the data points $\vec{z}_i$ are typically assumed to be independent and identically distributed (i.i.d.) according to an underlying (though unknown) probability distribution. Moreover, the model class $\modelclass$ is often parameterized, which means that each model $M \in \modelclass$ is uniquely identified by a parameter $\theta \in \Theta$ (in other words, there is a bijection between the model space $\modelclass$ and the parameter space $\Theta$).

In \emph{supervised learning}, the data space is split into an input (instance) space $\set{X}$ and an output space $\set{Y}$, that is,  $\set{Z} = \set{X} \times \set{Y}$. The interest, then, is to learn a mapping from $\set{X}$ to $\set{Y}$ that models, in one way or the other, the dependence of outputs (responses) on inputs (predictors); correspondingly, the model space $\modelclass$ typically consists of a class of such mappings. To this end, the learning algorithm $\algo$ is given a set 
$$
\set{D} = \big\{ \, (\vec{x}_i , y_i) \, \big\}_{i=1}^N \, \in \, (\set{X}\times\set{Y})^N
$$
of \emph{training examples} $\vec{z}_i =(\vec{x}_i , y_i) \in \set{X} \times \set{Y}$ as input. Important special cases of this setting include \emph{classification}, where $\set{Y}$ is a finite (usually small) set comprised of $K$ classes $\{ \lambda_1,  \ldots , \lambda_K\}$, and \emph{regression}, where outputs are real numbers ($\set{Y} = \mathbb{R}$).

In this paper, we are interested in the case where observations are imprecise and, therefore, characterized in terms of set-valued or fuzzy set-valued data. Subsequently, we therefore assume that, instead of precise data, the observations are given in the form of a sample of fuzzy data 
\begin{equation}\label{eq:fuzzydata}
\mathbb{D} = \big\{ \,  Z_i  \, \big\}_{i=1}^N \, \in \, \mathbb{F}(\set{Z})^N \enspace ,
\end{equation}
where $\mathbb{F}(\set{Z})$ is the set of all fuzzy subsets of the underlying data space $\set{Z}$. 

We like to emphasize that, in this setting, a fuzzy set $Z_i$ is supposed to represent information about an \emph{observation}, not about any kind of underlying ``true'' value or distribution; correspondingly, the specification of $Z_i$ will typically not involve any kind of statistical inference. In particular, our setting is completely coherent with the common statistical view of a data point $\vec{z}_i$ as the realization of a random variable characterized by a probability distribution, for example a normal distribution $\mathcal{N}(\bar{\vec{z}}, \sigma)$ with mean $\bar{\vec{z}}$ and standard deviation $\sigma$. Then, $Z_i$ would represent knowledge about the realization $\vec{z}_i$ and not about its expectation $\bar{\vec{z}}$.

\begin{table}
\begin{center}
\begin{tabular}{ll}
\hline
notation & meaning  \\
\hline
$\vec{z}_i$, $(\vec{x}_i , y_i)$    & (precise) data point, input/output sample \\
$\hat{\vec{z}}_i$, $\hat{y}_i$ & (precise) prediction/estimator \\
$Z_i, X_i, Y_i$ & sets or fuzzy sets (imprecise/fuzzy data)\\
\hline
$\mathcal{Z}$, $\mathcal{X}, \mathcal{Y}$ & data space, input space, output space \\
$\mathbb{F}(\mathcal{Z})$ & class of fuzzy subsets of $\mathcal{Z}$  \\
$\set{D}$, $\mathbf{D}$   & sample of (precise) data points, class of potential samples\\
$\mathbb{D}$   & sample of imprecise/fuzzy data\\
$\operatorname{INS}(\mathbb{D})$  & set of instantiations of $\mathbb{D}$\\
\hline
$L(y, \hat{y})$ & loss function, loss caused by $\hat{y}$ when compared to $y$ \\
$\mathcal{L}(Y, \hat{y} )$ & loss caused by $\hat{y}$ when compared to set $Y$\\
$\mathbb{L}( Y , \hat{y})$ & loss caused by $\hat{y}$ when compared to fuzzy set $Y$ \\ 
$\model$, $\modelclass$ & model, model space \\
$\mathcal{R}$, $\mathcal{R}_{emp}$ & risk, empirical risk \\
$\overline{\set{R}}_{emp}$ & aggregated (empirical) risk \\ 
$r_M$ & risk function mapping levels $\alpha \in (0,1]$ to risk values \\ 
\hline
\end{tabular}
\caption{Summary of the main notation used throughout the paper.}
\label{tab:notation}
\end{center} 
\end{table}

\section{Data Disambiguation}
\label{sec:extension}

Given a learning algorithm $\algo$ for precise data, the most straightforward approach to handling a fuzzy sample (\ref{eq:fuzzydata}) is to apply the well-known extension principle \cite{zade_tc75} to the mapping (\ref{eq:amap}). More formally, we define an \emph{instantiation} of the fuzzy sample (\ref{eq:fuzzydata}) as a sample 
$$
\set{D} = \big\{ \,  \vec{z}_i  \, \big\}_{i=1}^N 
$$
of precise data points, where $\vec{z}_i \in Z_i$ for all $i \in [N]=\{1, \ldots , N \}$. The degree of membership of $\set{D}$ in the fuzzy set of instantiations is given by
$$
\mu(\set{D}) = \min \Big\{ \mu_{Z_i} (\vec{z}_i) \with  i \in [N] \Big\} \enspace ,
$$
with $\mu_{Z_i} (\vec{z}_i)$ the degree of membership of $\vec{z}_i$ in $Z_i$. Then, according to the extension principle, the result of applying $\algo$ to the fuzzy data (\ref{eq:fuzzydata}) is a fuzzy set of models in $\modelclass$, with the degree of membership of $\model \in \modelclass$ given by
\begin{equation}\label{eq:ext}
\mu(\model) = \sup \Big\{ \, \mu(\set{D}) \with \algo(\set{D}) = \model \, \Big\} 
\enspace .
\end{equation}
We argue, however, that the application of the extension principle is not very meaningful in the context of learning from data. To ease the explanation for our reservations, let us consider the special case where the imprecise data is set-valued, i.e., the $Z_i$ are sets instead of fuzzy sets; as will be seen, our arguments obviously apply (``level-wise'') to the more general fuzzy case in exactly the same way. If data is set-valued, then the extension principle simply yields a subset of models from $\modelclass$, namely
\begin{equation}\label{eq:e1}
\mathcal{M} = \bigcup_{\set{D} \in \operatorname{INS}(\mathbb{D})} \algo(\set{D}) 
 \, \subseteq \, \modelclass \enspace ,
\end{equation}
where $\operatorname{INS}(\mathbb{D})$ is the (crisp) set of instantiations of $\mathbb{D}$.

Now, according to (\ref{eq:e1}), all instantiations are treated as equal, in the sense that each instantiation contributes a possible model and all the models thus produced are seen as equally plausible candidates. While this equal treatment of all instantiations is reasonable in common applications of the extension principle, where the variables of the function to be extended do not interact with each other, it can be questioned in the context of learning from data: A method inducing a model from a set of data always comes with certain \emph{model assumptions}, and under these assumptions, specific selections may appear more plausible than others! Or, stated differently, the underlying model assumptions introduce an implicit \emph{dependency} between the data points $\vec{z}_i \in Z_i$. This dependency, however, is ignored by the extension principle, which simply selects the $\vec{z}_i$ independently of each other.

This point is best explained by means of a simple example. Consider the problem of learning a regression function $M: \, \mathbb{R} \rightarrow \mathbb{R}$ from observations of the form $\vec{z}_i = (x_i , y_i) \in \mathbb{R}^2$. More specifically, suppose that the observed outputs are imprecise and therefore modeled as intervals $Y_i \subseteq \mathbb{R}$ (whereas the inputs $x_i$ are precise). Our learning algorithm $\algo$ assumes a linear dependency (i.e., the model space is given by $\modelclass = \{ x \mapsto \alpha + \beta \cdot x \with \alpha , \beta \in \mathbb{R} \}$) and fits the intercept $\alpha$ and the slope $\beta$ of the regression line using the method of least squares.

\begin{figure}
\begin{center}
\includegraphics[scale=0.6]{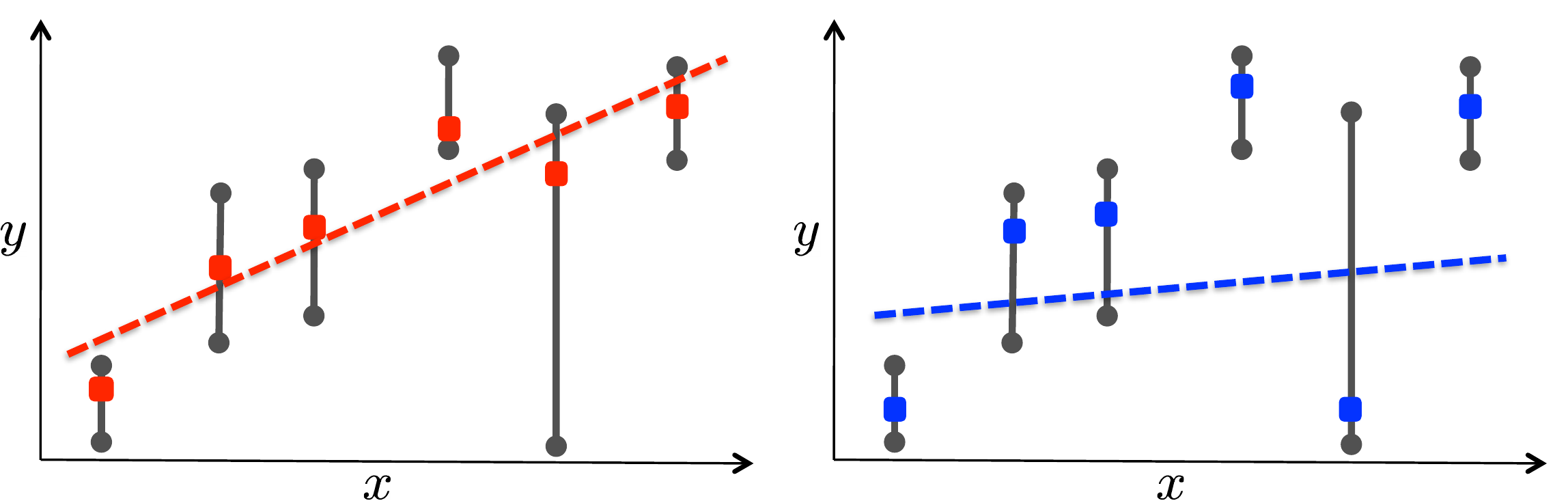}
\caption{Fit of a regression line for two different instantiations (indicated as dots) of the same interval-valued observations.}
\label{fig1}
\end{center}
\end{figure}

Figure \ref{fig1} shows a concrete example with two different instantiations of the same set-valued data and the corresponding regression lines. In this case, the first data/model combination (left picture) is arguably more plausible than the second one (right picture), simply because the first instantiation allows for a much better fit than the second one. In fact, the first instantiation is much more in agreement with the assumption of a linear relationship between inputs and outputs than the second one. Consequently, we argue that the first regression line should be considered as more plausible than the second one, at least in light of our assumption of a linear dependency. According to (\ref{eq:e1}) and the extension principle (\ref{eq:ext}), however, there is no difference between the two models.

\begin{figure}
\begin{center}
\includegraphics[scale=0.6]{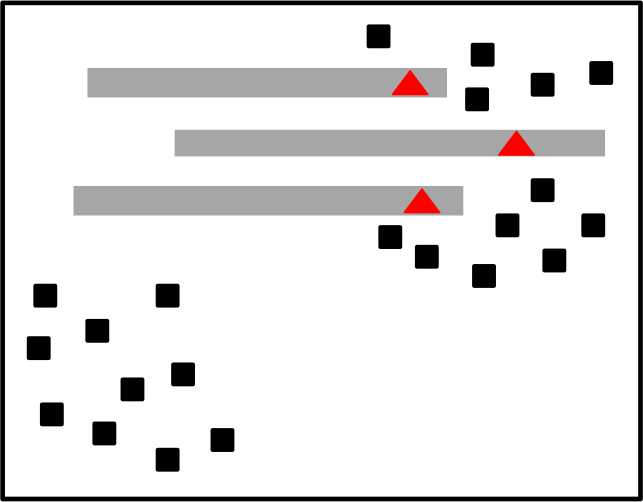} \hspace*{5mm}
\includegraphics[scale=0.6]{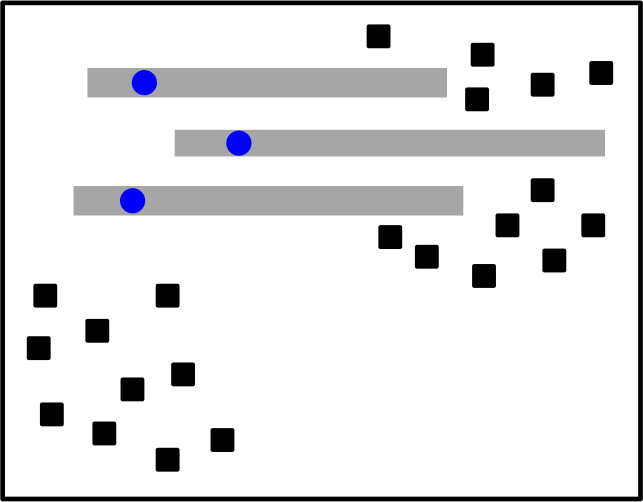}
\caption{Clustering of (partly) imprecise data in $\mathbb{R}^2$: The left instantiation appears more plausible than the right one.}
\label{fig:cluster}
\end{center}
\end{figure}

Another example is shown in Figure \ref{fig:cluster}, where the problem is to cluster data points $\vec{z} = (x,y) \in \mathbb{R}^2$. For three of the observations, the $x$-value is not known precisely and only characterized in terms of an interval; these observations are shown as grey rectangles in the picture. Now, assuming that the data is indeed well separated into subgroups, the instantiation in the left figure (red triangles) is arguably more plausible than the one in the right figure (blue circles). In fact, while the first instantiation allows for inducing a simple structure with two well-formed clusters, the second would imply a much less convenient structure.

What these examples show is that, in the context of learning from data, not only the data is providing information about the (unknown) model, but also the other way around: Against the background of the model assumptions underlying the model class $\modelclass$ and learning algorithm $\algo$, some instantiations of the imprecise or ambiguous data appear to be more plausible than others. Exploiting this insight in order to differentiate between more and less plausible instantiations is something that we refer to as \emph{data disambiguation} \cite{mpub136}. In other words, we consider an extension of standard model induction, in which we are not only interested in inferring properties of the data generating process, but also of the imprecisely observed data. Or, stated differently, we are not only interested in learning about the model given the data, but in learning about the model and the data simultaneously. 

As an aside, we note that, just like in standard statistics and machine learning, our approach takes the underlying model assumptions for granted and does not question them. Thus, model induction should be seen as a kind of \emph{conditional} inference, making hypothetical claims about the data generating process (and in our case even about the observed data itself) \emph{given the validity of the underlying model assumptions}. Needless to say, these assumptions are not always correct and, therefore, are often adapted or corrected by a data analyst if they seem to be incoherent with the data. The corresponding search for a proper model class, however, is outside the model induction process itself.

\section{A Loss Minimization Approach}

How can model induction be combined with data disambiguation? Here, we propose an approach based on the notion of (direct) loss minimization. Roughly speaking, instead of generalizing the learning algorithm, as done by the extension principle, we ``fuzzify'' an underlying loss function to be minimized by this algorithm. Thus, instead of fixing an instantiation first and fitting a model to this data afterward, we look for an optimal instantiation given a model; the model itself is then evaluated on the basis of this instantiation. 

In supervised learning, the main goal is typically to find a model $\model \in \modelclass$ with minimal \emph{risk}, that is, expected loss
\begin{equation}\label{eq:risk}
\mathcal{R}(M) \, = \, 
\int L \big( y , M(\vec{x}) \big) \, d \, \Prob(\vec{x}, y) \enspace ,
\end{equation}
where $L:\, \mathcal{Y} \times \mathcal{Y} \rightarrow \mathbb{R}$ is a loss function: For an input $\vec{x} \in \mathcal{X}$, this function compares the prediction $\hat{y} = M(\vec{x})$ with the true output $y$ and quantifies a corresponding penalty in terms of $L(y, \hat{y})$. Roughly speaking, the risk is a weighted average of these losses, with each input/output tuple $(\vec{x}, y)$ weighted according to its probability of occurrence. Thus, a risk minimizer 
$$
M^*  \in  \operatorname{arg}\min_{M \in \modelclass} \mathcal{R}(M)
$$
is a model that, on average, performs well in terms of the loss $L$. 

Obviously, the risk of a model $M$ cannot be computed directly, since the probability measure $\Prob$ in (\ref{eq:risk}), which specifies the data generating process, is unknown. What is often minimized as a substitute, therefore, is the \emph{empirical risk}
\begin{equation}\label{eq:erisk}
\mathcal{R}_{emp}(M) \, = \, \frac{1}{N}
\sum_{i=1}^N L \big( y_i , M(\vec{x}_i) \big)  \enspace ,
\end{equation}
i.e., the average loss on the training data $\set{D} = \left\{ (\vec{x}_i , y_i) \right\}_{i=1}^N$. Or, in order to avoid the problem of possibly \emph{overfitting} the data, a \emph{regularized} version of (\ref{eq:erisk}) is minimized:
\begin{equation}\label{eq:rerisk}
\mathcal{R}_{reg}(M) \, = \, \frac{1}{N}
\sum_{i=1}^N L \big( y_i , M(\vec{x}_i) \big)  + \lambda C(M) \enspace ,
\end{equation}
where $C(M)$ is a measure of the complexity of the model $M$ and $\lambda$ is a regularization parameter. In the following, we shall mostly stick to (\ref{eq:erisk}), keeping in mind that an extension to the regularized version (\ref{eq:rerisk}) can be realized in a rather straightforward way. 

\subsection{The Case of Set-Valued Data}

Again, for the ease of exposition, we consider the set-valued case first, before turning to the more general fuzzy case; moreover, we consider imprecision only for the output part while the inputs are supposed to be precise. 

Consider a candidate model $M$ and an imprecise observation $(\vec{x} , Y)$. With $\hat{y} = M(\vec{x})$, the set of possible losses of $M$ on this observation is then given by
$$
\big\{ \, L(y, \hat{y} ) \given y \in Y \, \big\}  \enspace .
$$
In agreement with the idea of data disambiguation, we should look at the smallest of these losses, namely
\begin{equation}\label{eq:gloss}
\mathcal{L}(Y, \hat{y} ) \, = \, \min \big\{ \, L( y, \hat{y} ) \given y \in Y \, \big\} \enspace ,
\end{equation}
and the value for which it is obtained:\footnote{We assume that $Y$ is closed and the minimum exists.}
$$
y^* \, = \, \operatorname{arg} \min \big\{ \, L( y, \hat{y} ) \given y \in Y \, \big\} \enspace .
$$
Given the model $M$, this value appears to be the most plausible in $Y$. 

The function $\mathcal{L}$ as defined in (\ref{eq:gloss}) can be seen as a generalized loss function, which, instead of comparing a (precise) prediction with a precise observation, compares a (precise) prediction with an imprecise (set-valued) observation. On the basis of this loss function, we can also generalize the empirical risk (\ref{eq:erisk}): 
\begin{equation}\label{eq:gerisk}
\mathcal{R}_{emp}(M) \, = \, \frac{1}{N}
\sum_{i=1}^N \mathcal{L} \big( Y_i , M(\vec{x}_i) \big)  \enspace .
\end{equation}
A minimizer 
\begin{equation}\label{eq:grm}
M^*  \in \operatorname{arg}\min_{M \in \modelclass} \mathcal{R}_{emp}(M)
\end{equation}
of this risk (or, alternatively, a regularized version thereof) is an optimal model and, at the same time, suggests a disambiguation of the data: For each imprecise observation $Y_i$, the most plausible precise value is
\begin{equation}\label{eq:disamb}
y_i^* \, = \, \operatorname{arg} \min \big\{ \, L( y_i, M^*(\vec{x}_i) ) 
\given y \in Y_i \, \big\} \enspace .
\end{equation}
Thus, the minimization of (\ref{eq:gerisk}) serves our original purpose and solves two problems simultaneously, namely the induction of a plausible model (\ref{eq:grm}) and a plausible disambiguation of the data (\ref{eq:disamb}).\footnote{This approach is connected to the ``minimin'' strategy for model selection under imprecision as proposed in \cite{utki_iv11}.}

So far, we have assumed that only the output value is imprecise, while the input values are precisely observed. Obviously, the whole approach can be generalized quite easily to the case of imprecise observations of the form $(X,Y) \subseteq \set{X} \times \set{Y}$. To this end, the loss function (\ref{eq:gloss}) is further generalized as follows:
\begin{equation}\label{eq:ggloss}
\mathcal{L}(M, X, Y) \, = \, \min \big\{ \, L( y, M(\vec{x}) ) \given (\vec{x},y) \in X \times Y \, \big\} \enspace .
\end{equation}

\subsection{The Case of Fuzzy Data}

In the set-valued case, each candidate model $M$ is evaluated in terms of a generalized empirical risk, that is, a risk function based on a generalized loss. This evaluation can be expressed equivalently in terms of a standard empirical risk on a properly selected (instantiated) data sample:
\begin{equation}\label{eq:e5}
\mathcal{R}_{emp}(M) \, = \, \frac{1}{N}
\sum_{i=1}^N L \Big( y_i^M , M \left(\vec{x}_i^M \right) \Big)  \enspace  ,
\end{equation} 
where 
\begin{align}\label{eq:sel}
\left(\vec{x}_i^M , y_i^M \right) & = \operatorname{SEL}(X_i, Y_i, M) \\
& = \operatorname{arg}\min \big\{ \, L( y_i, M(\vec{x}_i) ) \given (\vec{x}_i,y_i) \in X_i \times Y_i \, \big\} \nonumber
\end{align}
is the disambiguation of $(X_i, Y_i)$ under $M$. A best model 
\begin{equation}\label{bestmodel}
M^* \, = \, \operatorname{arg} \min_{M \in \modelclass}  \mathcal{R}_{emp}(M)
\enspace ,
\end{equation}
supposed to be unique here, is then chosen, which in turn leads to a unique disambiguation 
$$
\Big\{ \left( \vec{x}_i^{M^*} , y_i^{M^*} \right) \Big\}_{i=1}^N
$$
of the original (imprecise) data.
In the more general case of fuzzy data, the same approach can be realized \emph{level-wise}, i.e., for each level-cut 
$$
\Big\{ \left( [X_i]_{\alpha}, [Y_i]_{\alpha} \right)  \Big\}_{i=1}^N
$$
of the fuzzy data 
$$
\big\{ ( X_i , Y_i ) \big\}_{i=1}^N \subseteq  \mathbb{F}(\set{X}) \times \mathbb{F}(\set{Y})  \enspace .
$$ 
Then, for a fixed model $M$, data disambiguation does not yield a unique selection (\ref{eq:sel}), but instead a potentially different selection for each level cut. In other words, the selection is now a mapping 
$$
\alpha \mapsto  
\left(\vec{x}_i^M(\alpha) , y_i^M(\alpha) \right) 
= \operatorname{arg}\min \big\{ \, L( y_i, M(\vec{x}_i) ) \given (\vec{x}_i,y_i) \in [X_i]_{\alpha} \times [Y_i]_{\alpha} \, \big\} .
$$
In \cite{dubo_ge08}, a mapping of that type is called a \emph{gradual element} (in a fuzzy set). Likewise, a mapping from levels to (empirical) risk values can be associated with each model $M$:
\begin{equation}\label{eq:riskfunction}
r_M:\, (0,1] \rightarrow \mathbb{R}, \,  \alpha \mapsto  
\frac{1}{N}
\sum_{i=1}^N L \Big( y_i^M(\alpha) , M \left(\vec{x}_i^M(\alpha)\right) \Big)
\end{equation}
Note that the risk function $r_M$ thus defined is non-decreasing.

\begin{figure}
\begin{center}
\hspace*{-9mm}
\includegraphics[scale=0.37]{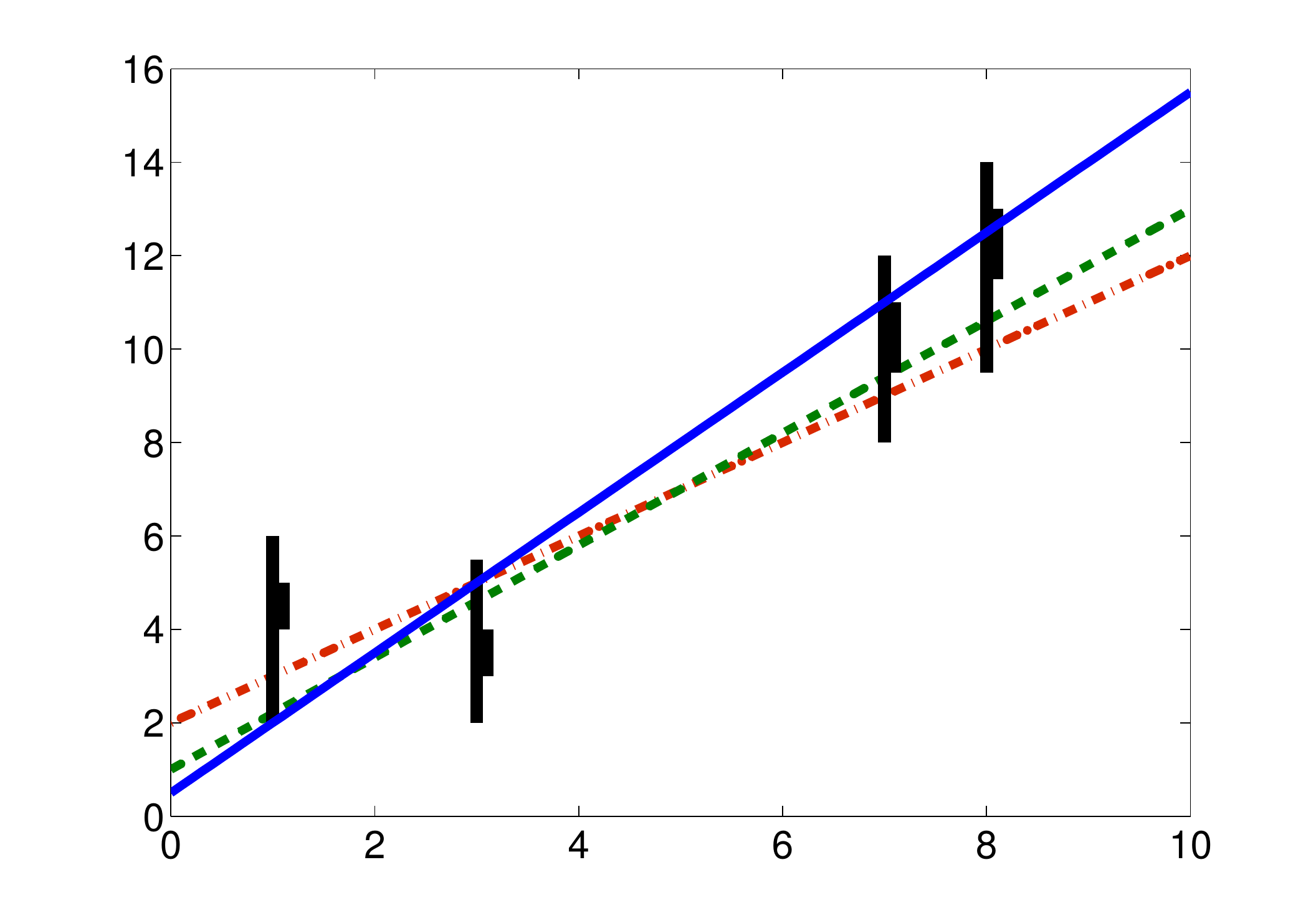} \hspace*{-6mm}
\includegraphics[scale=0.37]{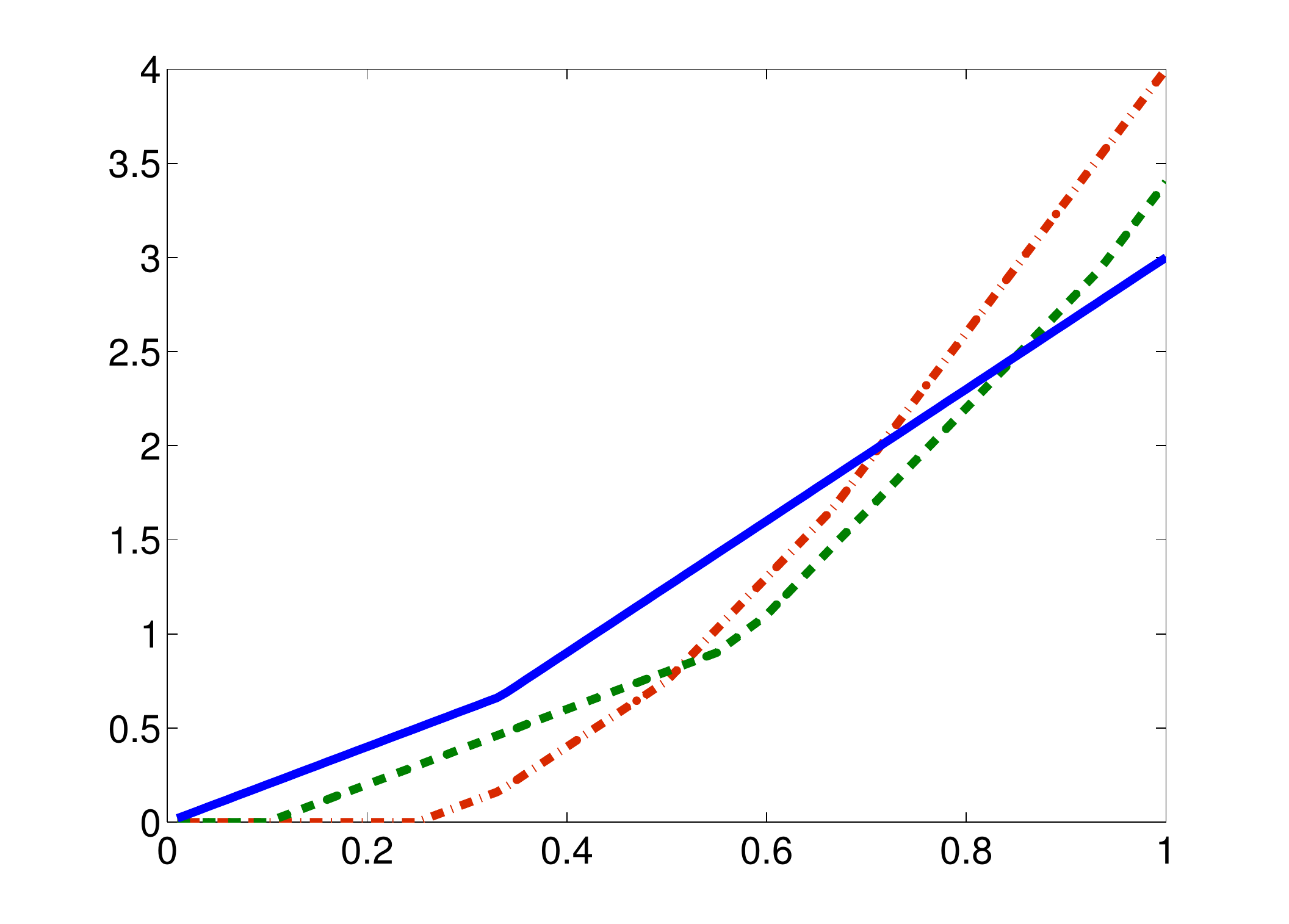}


\caption{Left: Fuzzy output data indicated by the support (thin line) and the core (thick line) of a trapezoidal fuzzy set; moreover, three regression lines approximating this data. Right: The corresponding risk functions $r_M(\cdot)$ in the same color and line style.}
\label{fig2}
\end{center}
\end{figure}

The problem of comparing models now comes down to comparing risk functions. This problem is non-trivial, since there is no natural total order on such functions. Obviously, a model $M$ is (weakly) preferred to another model $M'$, written $M \succeq M'$, if $r_M \leq  r_{M'}$, i.e., $r_M(\alpha) \leq r_{M'}(\alpha)$ for all $0 < \alpha \leq 1$. The relation $\succeq$ thus defined is only a partial order on the model class $\modelclass$, as models $M$ and $M'$ may also be incomparable (i.e., neither $M \succeq M'$ nor $M' \succeq M$). Figure \ref{fig2} shows a simple (one-dimensional) example for the case of regression, namely three regression lines approximating four observations with fuzzy output; all three models are incomparable amongst each other, that is, none of them dominates any other one in terms of the associated risk function.    

This situation can be handled in different ways. First, one may accept the non-uniqueness of the result, i.e., the existence of several (Pareto) optimal models; here, a model $M$ is optimal (non-dominated) if there is no model $M'$ such that $M' \succ M$, that is, $M' \succeq M$ and $M \not\succeq M'$.

Second, one may refine the partial oder $\succeq$ as defined above into a total order. For example, a model $M$ could be evaluated in terms of the \emph{aggregated} risk
\begin{equation}\label{eq:arisk}
\overline{\set{R}}_{emp}(M) \, = \, \int_0^1 r_M(\alpha) \, d \, \alpha \enspace ,
\end{equation}
and models could then be compared in terms of these values:
$$
\big( M \succeq M' \big) \, \Leftrightarrow \, \Big( \, 
\overline{\set{R}}_{emp}(M) \leq \overline{\set{R}}_{emp}(M') \, \Big)
$$
The model induction problem then comes down to finding a minimizer of (\ref{eq:arisk}):
\begin{equation}\label{bestfuzzymodel}
M^* \, \in \, 
\operatorname{arg} \min_{M \in \modelclass} \overline{\set{R}}_{emp}(M)
\end{equation}
Interestingly, by exchanging summation and integration, (\ref{eq:arisk}) can also be written as a standard (empirical) risk with a modified loss function:
\begin{equation}\label{eq:arisk2}
\overline{\set{R}}_{emp}(M) \, = \, \frac{1}{N} \sum_{i=1}^N 
\mathbb{L} \big( Y_i , M(\vec{x}_i) \big) \enspace ,
\end{equation}
where
\begin{equation}\label{eq:fuzzyloss}
\mathbb{L} \big( Y , \hat{y} \big) \, = \, 
\int_0^1 \mathcal{L} \big( [Y]_{\alpha} , \hat{y} \big) \, d \, \alpha  
\end{equation}
is a ``fuzzy'' loss function that compares a (precise) prediction with a fuzzy set-valued observation. 

Expression (\ref{eq:arisk2}) holds in the case of precise input and fuzzy output data but needs to be generalized further if input data is fuzzy, too.

\subsection{Fuzzy Losses for Regression}

\begin{figure}
\begin{center}
\includegraphics[scale=0.35]{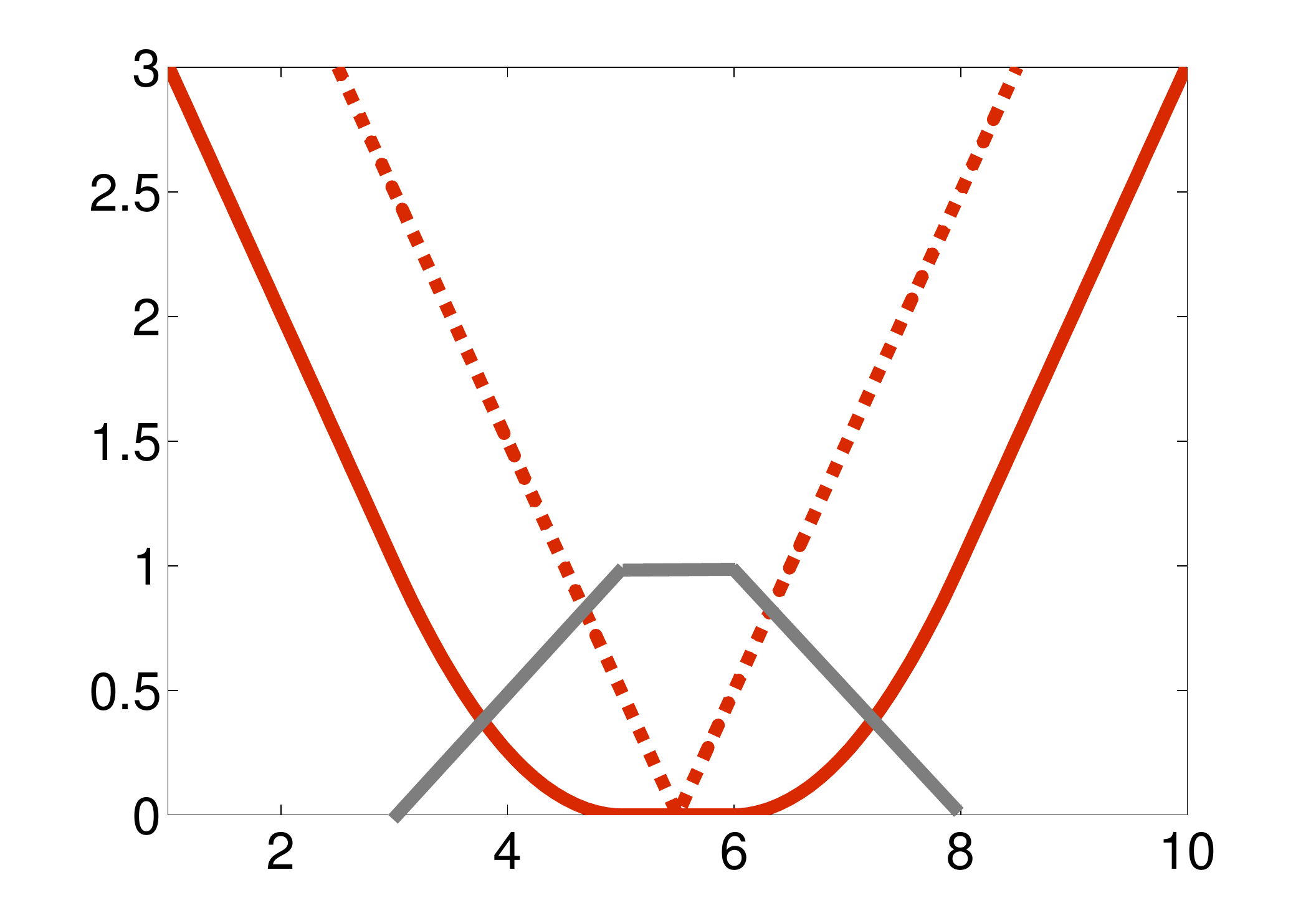} \hspace*{-10mm}
\includegraphics[scale=0.35]{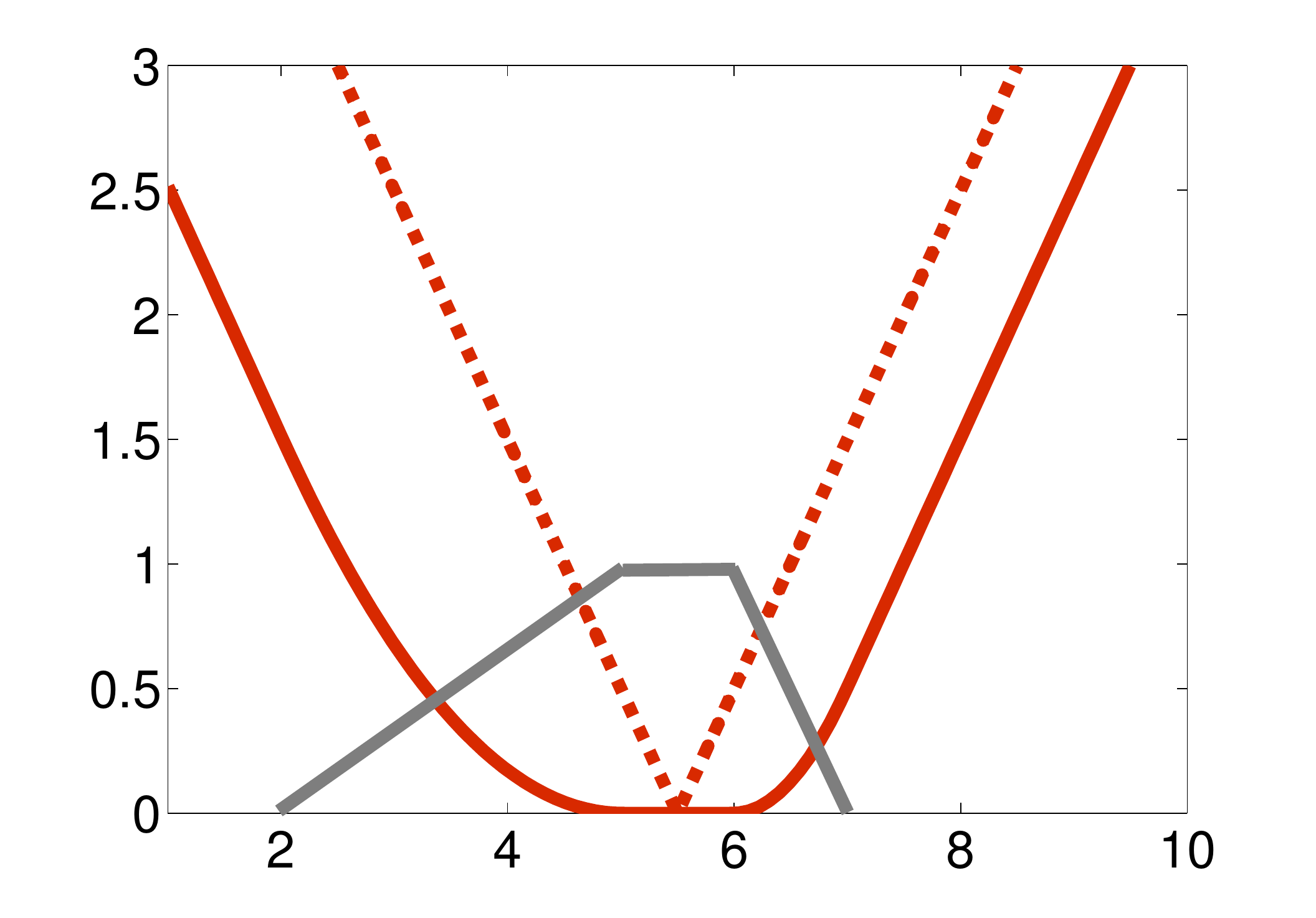}

\caption{Left: Example a fuzzy loss function $\hat{y} \mapsto  \mathbb{L}(Y, \hat{y})$, where $Y$ is the trapezoidal fuzzy set shown in grey, and $L$ is the $L_1$ loss. Right: The same function in the case of an asymmetric fuzzy set.}
\label{fig3}
\end{center}
\end{figure}

The fuzzy loss function (\ref{eq:fuzzyloss}) compares a fuzzy value $Y$ with a (predicted) precise value $\hat{y}$. An example of such a loss is shown in Figure \ref{fig3} for the case of regression. More specifically, this function is a fuzzy version of the absolute ($L_1$) loss 
$$
L(y,\hat{y})= |y-\hat{y}| \enspace , 
$$
which is shown as a dashed line (as a function of $\hat{y}$ for fixed $y=5.5$). The fuzzy loss (solid line) is given by the map $\hat{y} \mapsto  \mathbb{L}(Y, \hat{y})$, where $Y$ is the trapezoidal fuzzy set shown in grey.

Interestingly, a fuzzification of the $L_1$ loss based on a \emph{triangular} fuzzy set $Y$ with mid-point $y$ and support $(y - \delta , y + \delta)$ leads to a kind of Huber-loss \cite{hube_rs}:
$$
\mathbb{L} \big( Y , \hat{y} \big) \, = \, 
\left\{ \begin{array}{ll}
\frac{1}{2} (y - \hat{y})^2/\delta & \text{ if } \hat{y} \leq \delta \\
|y - \hat{y}| - \frac{1}{2} \delta & \text{ if } \hat{y} > \delta
\end{array} \right.
$$
This loss behaves like the quadratic ($L_2$) loss for small errors and like the $L_1$ loss for larger deviations. This kind of loss function is very popular in robust statistics, as it combines two interesting properties: Like the absolute error $L_1$, it is much less sensitive toward outliers than, for example, $L_2$, but at the same time, it avoids the non-differentiability of $L_1$.

As can be seen, our approach to learning from fuzzy data based on generalized loss functions includes methods such as M-estimation with Huber-loss as specific cases; methods for Huber M-estimation have been studied quite intensively in the literature \cite{mang_rl00}. It needs to be mentioned, however, that our approach is in a sense more general, especially as it allows for modeling each fuzzy value and, therefore, the corresponding loss function \emph{individually} instead of applying the same loss function to each observation (recall, for instance, the example of an asymmetric fuzzy loss in Figure \ref{fig3} (right)). In other words, our approach is \emph{sample-specific} in the sense that a specific loss function can be defined for each sample point. To make this more clear, we may also write the fuzzy loss (\ref{eq:fuzzyloss}) using a slightly different notation:
$$
\mathbb{L} \big( Y , \hat{y} \big) \, = \, \mathbb{L}_Y \big( y , \hat{y} \big)
\enspace ,
$$
where $y$ could be an observed value, and the fuzzy set $Y$ is used to specify a region of imprecision around this observation. Thus, the fuzzy loss $\mathbb{L}_Y$ is a standard loss function (defined on pairs of precise values) ``modulated'' by the fuzzy set $Y$ around $y$. 

Another important loss function we can mimic is the $\epsilon$-insensitive loss that plays an important role in support vector regression \cite{scho_lw}:
$$
L(y, \hat{y}) = 
\left\{ \begin{array}{cl}
0 & \text{ if } |y-\hat{y}|\leq \epsilon \\
|y - \hat{y}| - \epsilon & \text{ if } |y-\hat{y}| > \epsilon
\end{array} \right.
$$
This loss is obtained as a special case of (\ref{eq:fuzzyloss}) with $Y$ given by the interval $[y - \epsilon , y + \epsilon]$. 

The use of a trapezoidal fuzzy set $Y$ with core $[y - \epsilon , y + \epsilon]$ and support $[y - \delta , y + \delta]$ nicely combines the two types of loss discussed above: $\mathbb{L}_Y$ is insensitive in the core, behaves quadratically in the boundary region $(y-\delta , y-\epsilon) \cup (y +\epsilon, y + \delta)$ and like $L_1$ outside the support.

\subsection{Fuzzy Losses for Classification}

In classification problems, the output space $\set{Y}$ is a finite set comprised of $K$ classes $\{ \lambda_1,  \ldots , \lambda_K\}$. The most typical loss function is the 0/1 loss $L(y, \hat{y}) = \llbracket y \neq \hat{y} \rrbracket$. Now, suppose the output is characterized by a fuzzy subset $Y$ of $\set{Y}$, that is, by a membership degree $\mu_Y(\lambda_i)$ for each class label $\lambda_i \in Y$. The fuzzy loss function (\ref{eq:fuzzyloss}) is then given as follows:
$$
\mathbb{L}(Y, \hat{y}) = 1 - \mu_Y(\hat{y}) \enspace .
$$
Thus, the higher the membership degree of the predicted class $\hat{y}$, the smaller the loss. An interesting special case is obtained for a fuzzy set of the type
\begin{equation}\label{eq:sc}
\mu_Y(\lambda) \, = \, 
\left\{ \begin{array}{cl}
1 & \text{ if } \lambda = \lambda_k  \\
1-w & \text{ if } \lambda \neq \lambda_k
\end{array} \right. \enspace ,
\end{equation}
for some $k \in [K]$ and $w \in [0,1]$. Using this fuzzy set for modeling the observation of class label $\lambda_k$ corresponds to a \emph{discounting} of this observation: Although $\lambda_k$ is regarded as completely plausible, the other class labels are not fully excluded either; or, stated differently, $w$ can be seen as a degree of certainty that the observed class is indeed $\lambda_k$. For a fuzzy observation of that kind,
$$
\mathbb{L}(Y, \hat{y}) \, = \, 
\left\{ \begin{array}{cl}
0 & \text{ if } \hat{y} = \lambda_k  \\
w & \text{ if } \hat{y} \neq \lambda_k
\end{array} \right. \enspace ,
$$
which means that the penalty for a misclassification is effectively reduced from 1 to $w$. In other words, the training example is \emph{weighted} by the factor $w$. Again, learning from weighted examples (aka \emph{instance weighting}) has been studied intensively in the literature \cite{shim_ip00}.

An important class of loss functions in binary classification is the so-called \emph{margin losses} \cite{ross_mm03}. Instead of merely checking whether a prediction is on the right or the wrong side of the decision boundary, as the 0/1 loss does, such losses depend on \emph{how much} on the right or wrong side the prediction is.  By preferring ``very correct'' predictions to simply correct ones, they enforce a ``large margin'' between the classes, i.e., they tend to separate the classes as much as possible. 

More formally, let $\mathcal{Y} = \{ -1 , +1 \}$ encode the two classes (negative and positive), and suppose that $\modelclass$ is a class of \emph{scoring classifiers} $M:\, \mathcal{X} \rightarrow \mathbb{R}$; a positive score $s = M(\vec{x}) > 0$ suggests that $\vec{x}$ belongs to the positive class, whereas a negative score suggests that $\vec{x}$ is negative. A margin loss is a function of the form 
\begin{equation} \label{marginloss}
L(y, s) \, = \, f(y   s)  \enspace ,
\end{equation}
where $f: \, \mathbb{R} \rightarrow \mathbb{R}$ is non-increasing. Thus, a margin loss penalizes scores instead of binary predictions, and the larger (smaller) the score in the case of a positive (negative) class, the smaller the loss. Important examples of (\ref{marginloss}) include the hinge loss
\begin{equation} \label{hingeloss}
L(y, s) \, = \, f(y   s)  = \max\big( 1 - y s , 0 \big) 
\end{equation}
used in support vector machines \cite{vapn_sl98,scho_lw}, the exponential loss
\begin{equation} \label{exploss}
L(y, s) \, = \, f(y   s)  = \exp(- y s ) 
\end{equation}
used in boosting algorithms \cite{scha_ts90}, and the logistic loss
\begin{equation} \label{logloss}
L(y, s) \, = \, f(y   s)  = \log\big( 1 + \exp(- y s ) \big) 
\end{equation}
closely connected with logistic regression. 

\begin{figure}
\begin{center}

\includegraphics[scale=0.4]{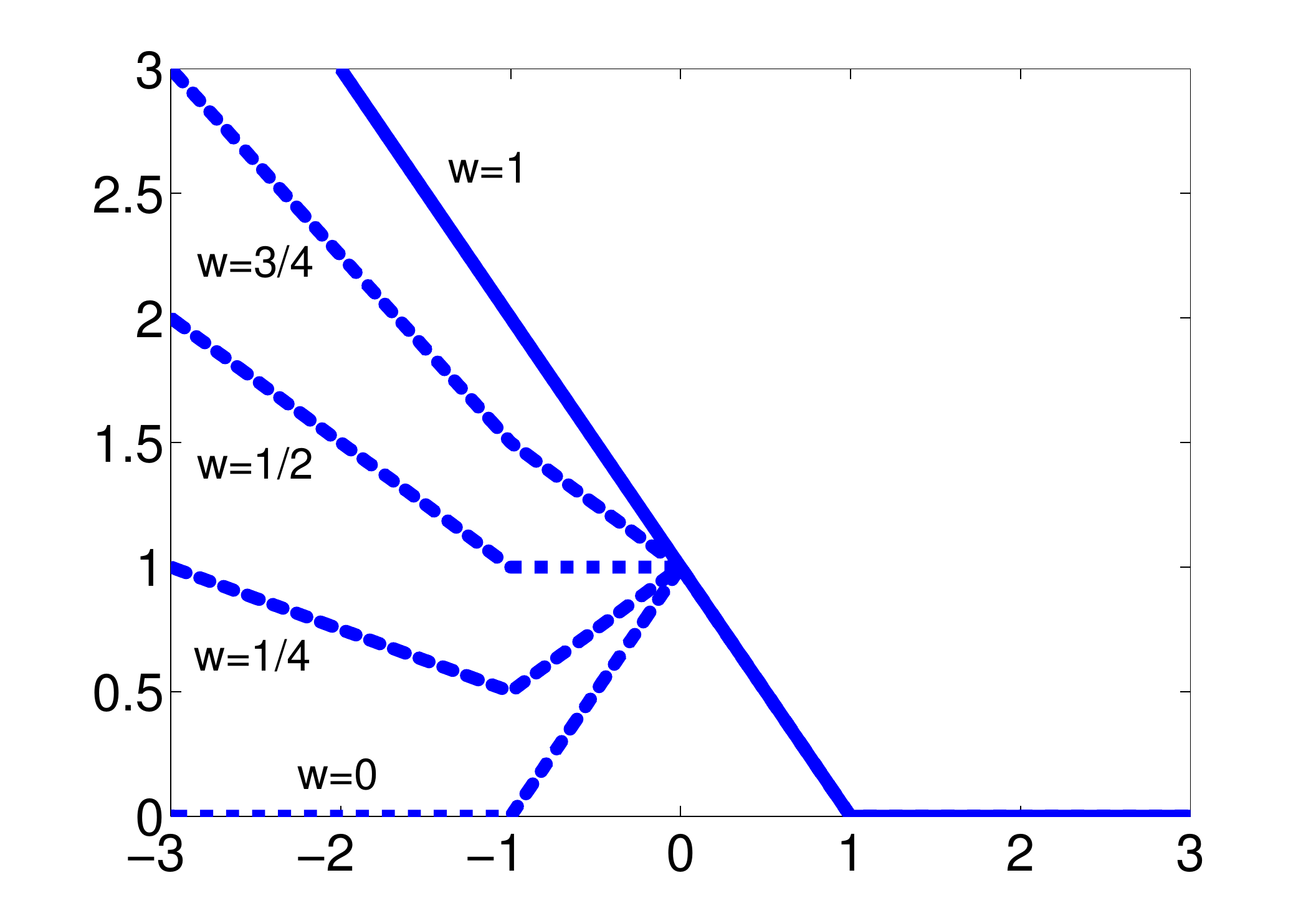} 

\includegraphics[scale=0.4]{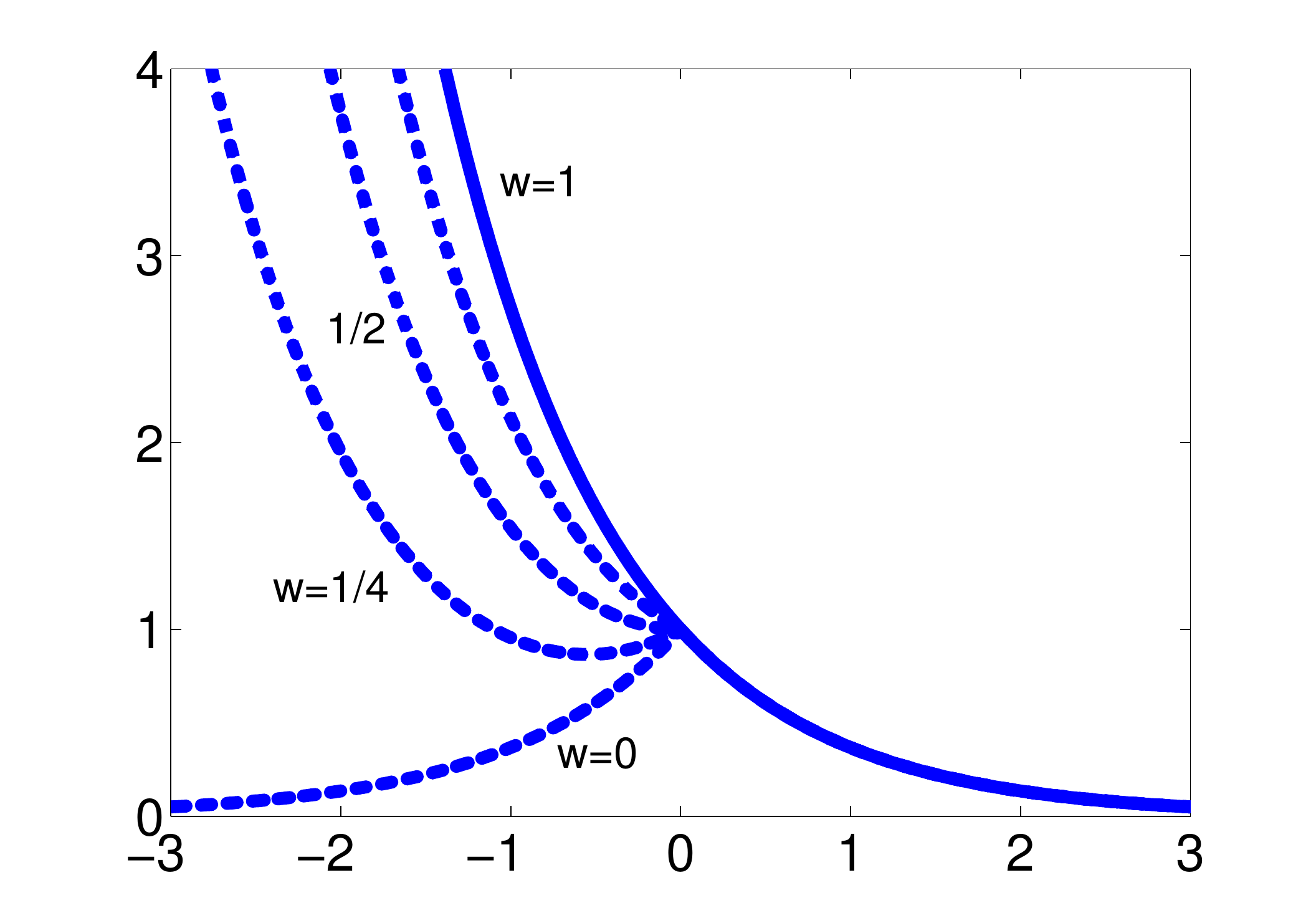}

\includegraphics[scale=0.4]{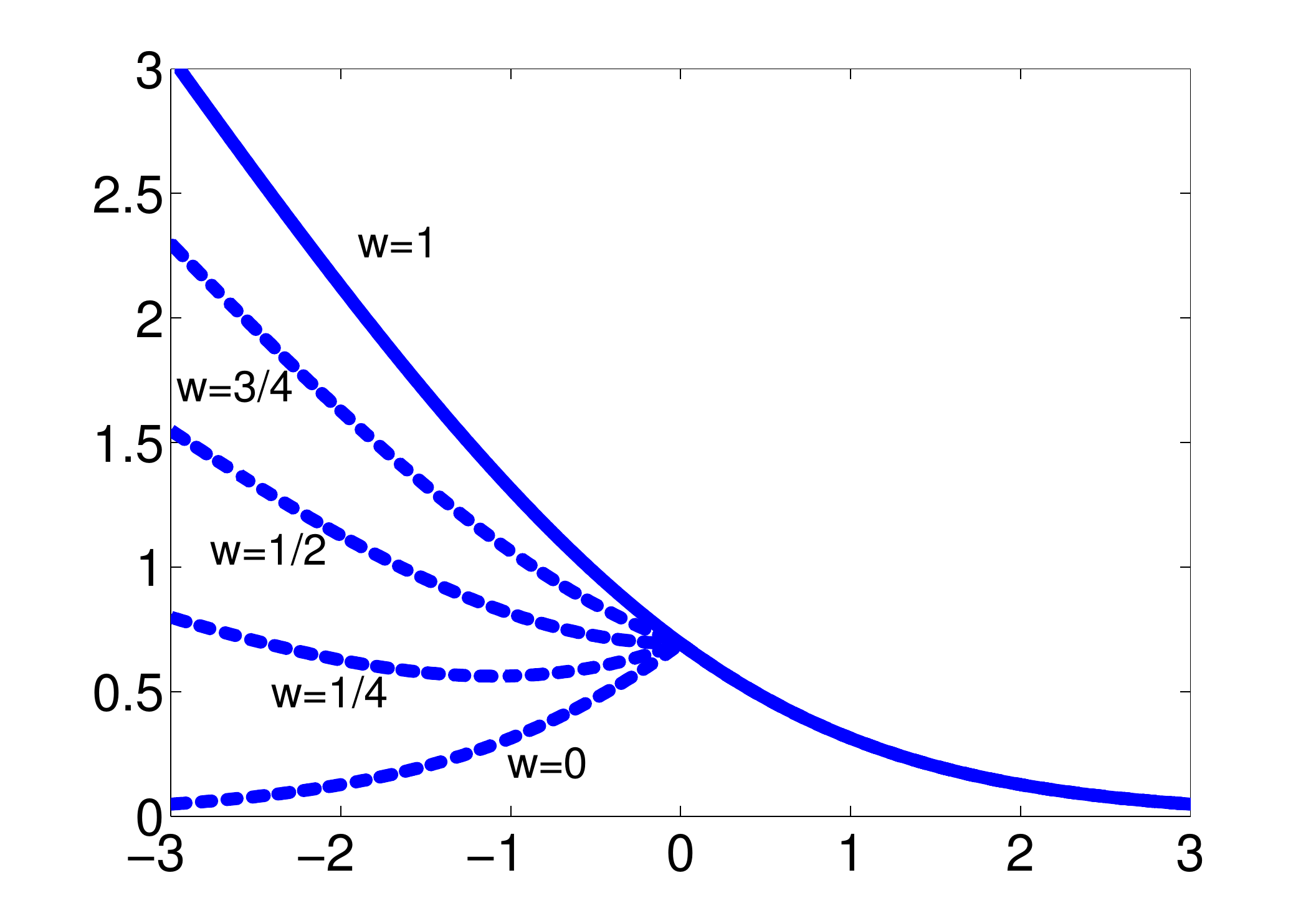}

\caption{Original margin losses (solid line) and discounted version (\ref{fmloss}) for different values of $w$: hinge (top), exponential (middle) and logistic (bottom).}

\label{fig:mlosses}
\end{center}
\end{figure}

Now, suppose again that the output is characterized by a fuzzy subset $Y$ of $\set{Y}$, that is, by a membership degrees $\mu_Y(-1)$ and $\mu_Y(+1)$ for the negative and positive class, respectively. More specifically, consider again the special case (\ref{eq:sc}):
\begin{equation}\label{eq:disc}
\mu_Y(\lambda) \, = \, 
\left\{ \begin{array}{cl}
1 & \text{ if } \lambda = y  \\
1-w & \text{ if } \lambda = \bar{y} 
\end{array} \right. \enspace ,
\end{equation}
where $\{ y, \bar{y} \} =  \{ -1, +1 \}$ and $w$ can be interpreted as a degree of confidence in $y$.  
Then, it is not difficult to show that the fuzzy loss function (\ref{eq:fuzzyloss}) is given by
\begin{equation}\label{fmloss}
\mathbb{L}(Y, s) = f_w(ys) = w \cdot f(y s) + (1-w) \cdot f( |ys| )  \enspace .
\end{equation}
Please note that $f_w$ coincides with the original margin loss $f$ if $ys > 0$, i.e., if the prediction $s = M(\vec{x})$ is in favor of the more likely class $y$; thus, the difference only concerns the negative part. Figure \ref{fig:mlosses} shows the graph of (\ref{fmloss}) for different margin losses and different values of $w$. 

As can be seen, the loss (\ref{fmloss}) looses the properties of monotonicity and convexity for sufficiently small values of $w$. Apart from the fact that this is certainly undesirable from a computational perspective, as it makes optimization more difficult, the non-monotone behavior of the loss may also be surprising at first sight. At second sight, however, it makes perfect sense. In fact, one has to keep in mind that, in contrast to the simple 0/1 loss, a margin loss pursues two goals at the same time, namely correct classification and separation of the data. To comply with the first goal, the penalty should decrease with decreasing $w$, just like in the case of the 0/1 loss; this is why $f_w \leq f_{w'}$ for $w \leq w'$. At the same time, however, an increase of the margin is rewarded. Taking both effects together, that is, a discounted penalty for misclassification and a reward for an increased margin, it is possible that an incorrect classification with a large margin in penalized less than a correct classification with a small margin.

Moreover, one should note that the fuzzy margin losses are fully in agreement with our idea of data disambiguation. This can be seen most clearly for $w=0$, which corresponds to the case where both labels, positive and negative, are considered completely plausible (in other words, no label information is given). Here, the loss is a symmetric function around 0: Putting an instance directly on the decision boundary, and thereby expressing maximal ambiguity, is the worst solution and penalized with the highest loss. The larger the distance from the decision boundary, regardless to what side, the smaller the loss becomes. Or, stated differently, the more pronounced the prediction in favor of one of the classes, the better it is.

\begin{figure}
\begin{center}

\includegraphics[scale=0.40]{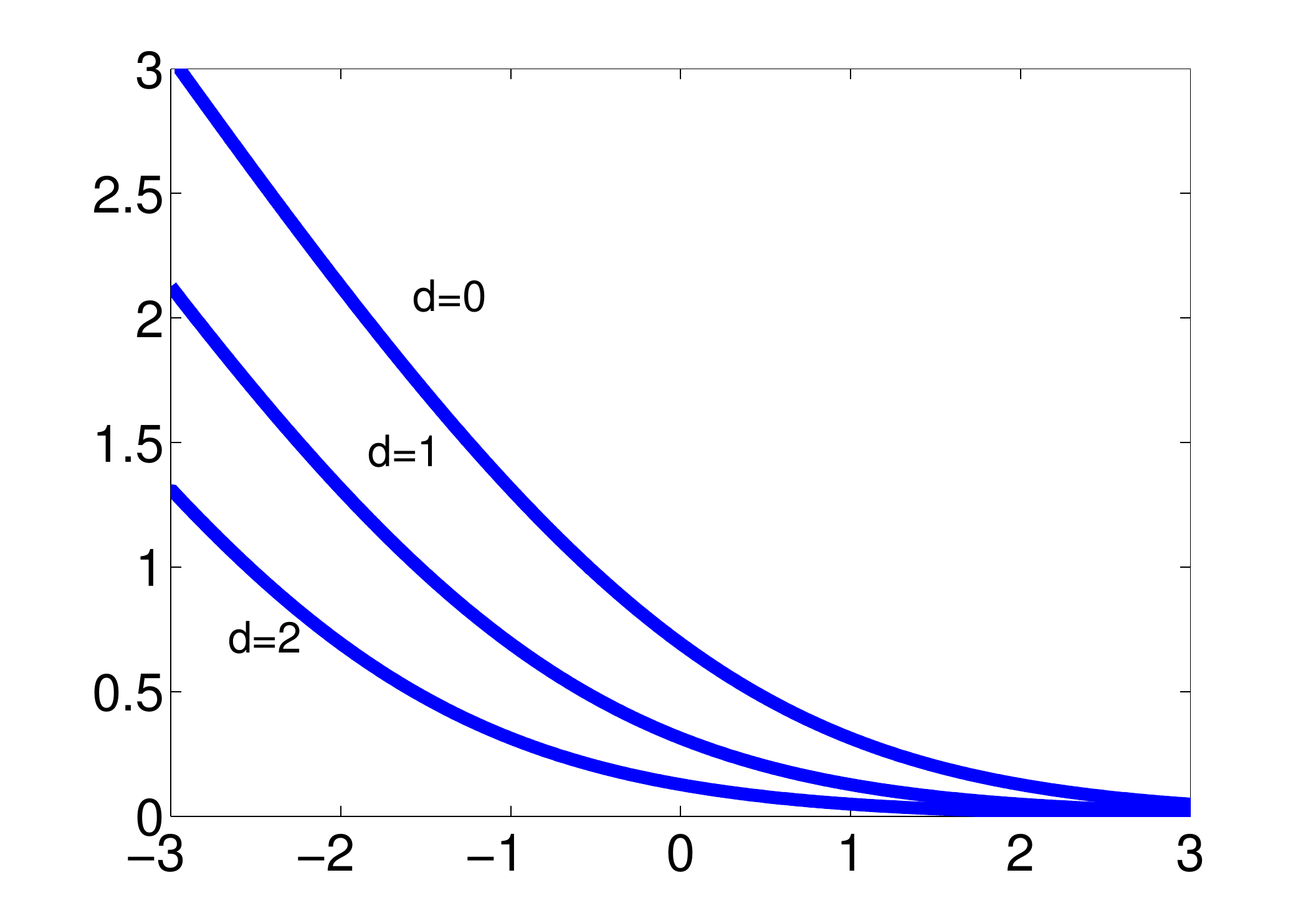} 

\caption{Fuzzy margin loss (\ref{eq:mloss2}) as a function of $s$ for different values of $d$, with $L$ the logistic loss function.}
\label{fig:mloss2}
\end{center}
\end{figure}

So far, we only considered imprecision of the dependent variables and assumed the predictor variables to be precise. Without going into detail, we note that the predictors can of course be affected by imprecision, too, and that the effect on the loss function is different in this case. For example, suppose that a predictor $\vec{x}$ is represented by a (closed) contiguous region $X \subseteq \set{X}$, such as a rectangle or a ball. The scores that can be produced for this instance by a model $M$ are then given in the form of an interval  
$$
\big[ \, \min\{ M(\vec{x}) \with \vec{x} \in X \} , \, 
\max\{ M(\vec{x}) \with \vec{x} \in X \} \, \big] \enspace ,
$$
which can also be written as $[s - d, s + d]$ with some $d \geq 0$ and $s$ the middle point. Applying the generic loss function (\ref{eq:ggloss}) to this case (with precise output $y$), and assuming $L$ to be a margin loss, we obtain
\begin{equation}\label{eq:mloss2}
\mathcal{L}(M,X, y) = L \big( \max\{ y (s-d), y (s+d) \} \big) \enspace .
\end{equation}
Thus, the loss function $L$ is ``shifted to the left'' by $d$ units; see Figure \ref{fig:mloss2} for an illustration.

\section{Comparison with Denoeux's Approach}

Denoeux addressed a quite similar problem in his recent articles \cite{deno_ml11,deno_ml11b}. More specifically, he addressed the problem of learning from imprecise data, represented in terms of fuzzy sets or belief functions, within a probabilistic framework and, for this purpose, proposed an extension of maximum likelihood inference. Without going into technical details, we shall try to highlight the main conceptual differences between Denoeux's approach (subsequently referred to as GMLI for Generalized Maximum Likelihood Inference) and ours, presenting ideas of the former in terms of our notation.\footnote{A special case of this approach was already introduced in \cite{come_lf09}.} 

Roughly speaking, given a sample of imprecise data $\mathbb{D} = \{  Z_i  \}_{i=1}^N$, 
Denoeux defines the plausibility of a model $M_\theta$ identified by a parameter $\theta$ in terms of a normalized likelihood; the likelihood of $\theta$ is in turn defined by the probability that the data-generating process specified by $\theta$ produces an instantiation $\set{D} = \{  \vec{z}_i  \}_{i=1}^N \in \operatorname{INS}(\mathbb{D})$:
$$
\pi(M) = \pi(\theta)  \propto \Prob( \set{D} \in \mathbb{D}  \given \theta ) =
\prod_{i=1}^N \Prob( Z_i \given \theta )
$$
This probability can also be written as
\begin{equation}\label{eq:avg}
\int_{\set{D} \in \operatorname{INS}(\mathbb{D})} 
\Prob( \set{D} \given \theta ) \, d \mu(\set{D})
=
\int_{\set{D} \in \operatorname{INS}(\mathbb{D})} 
\prod_{i=1}^N \Prob( \vec{z}_i \given \theta ) \, d \mu(\set{D})  \enspace ,
\end{equation}
where $\mu(\cdot)$ measures the plausibility of instantiations. 
This already reveals the most important difference between Denoeux's approach and ours: In the former, a model is evaluated, not by looking at how it fits the \emph{most favorable} instantiation of the imprecise data, but how it fits \emph{all possible instantiations} simultaneously. In fact, as can be seen from (\ref{eq:avg}), the score of a model is obtained by summing (averaging) its likelihood degrees (on precise samples) over all instantiations.

\begin{figure}
\begin{center}

\includegraphics[scale=0.40]{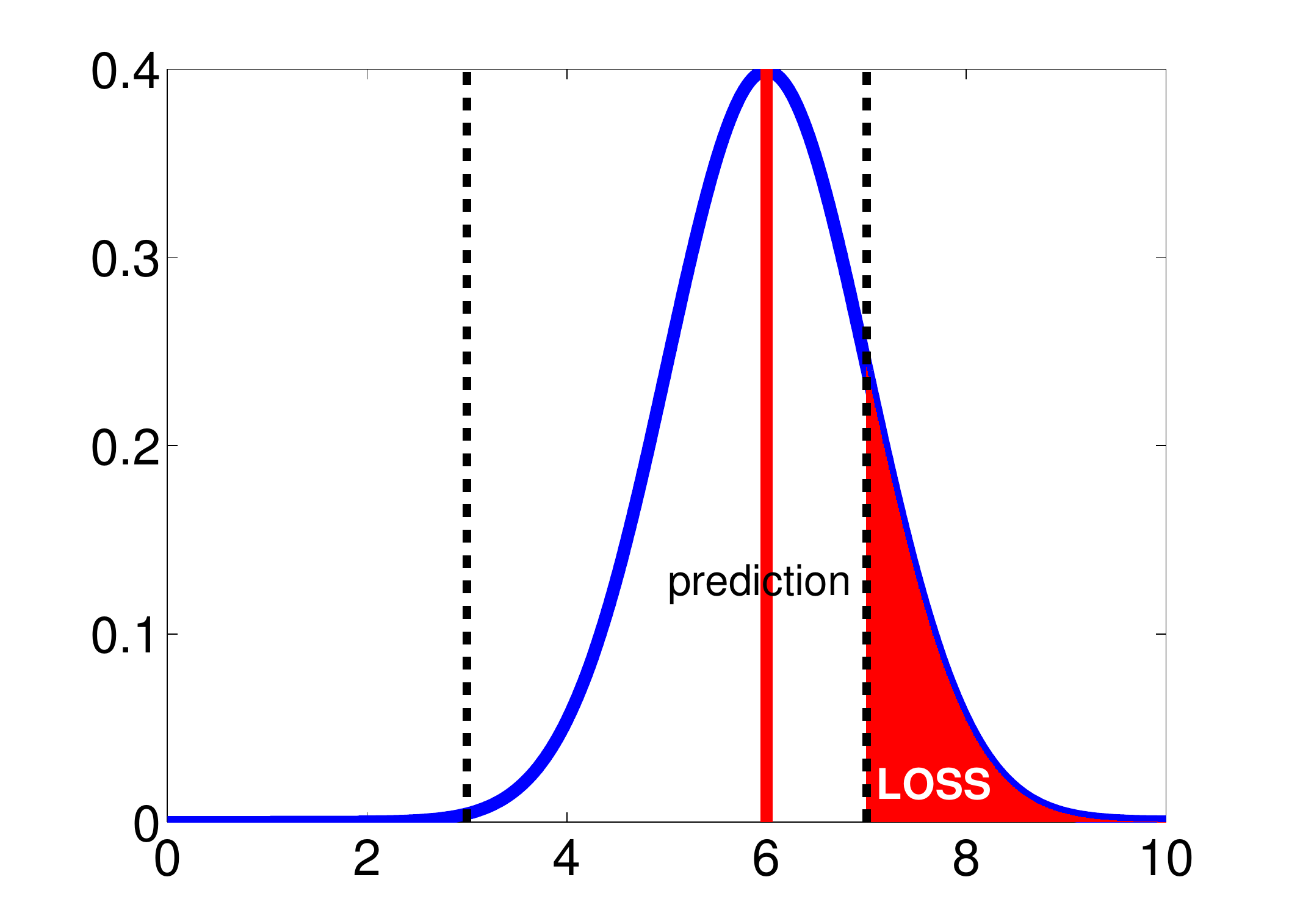} 

\vspace*{0mm}

\caption{Illustration of the loss caused by a prediction $\hat{y}_i = M(\vec{x}_i) = 6$: The loss is given by the negative logarithm of the area outside the observed interval $[3,7]$, which essentially corresponds to the shaded area on the right side.}
\label{fig:pic-den-1}
\end{center}
\end{figure}

The difference may perhaps become even more clear when looking at GMLI from our loss minimization perspective. As already mentioned earlier, likelihood maximization and loss minimization are closely connected, and maximizing the log-likelihood can typically be considered as minimizing an additive loss on the training data, namely the log-loss. As an illustration, consider the simple case of (one-dimensional) regression, where the observed response is supposed to follow a normal distribution. Thus, given (precise) training data $\set{D} = \{  (\vec{x}_i, y_i)  \}_{i=1}^N$, the likelihood is of the form
\begin{equation}\label{eq:gaussian}
c \prod_{i=1}^N \frac{1}{\sigma} \exp \left( - \frac{1}{2} \left( 
\frac{M(\vec{x}_i) - y_i}{\sigma} \right)^2 \right) \enspace ,
\end{equation}
where $c$ is a normalizing constant, and the minimizer of the logarithm of that likelihood is obviously equivalent to the least squares estimator
$$
M^* = \operatorname{arg}\min \sum_{i=1}^N \big( M(\vec{x}_i) - y_i \big)^2 \enspace .
$$
In the case where a response $y_i$ is imprecise, the contribution to the likelihood is a factor of the form $\Prob( M(\vec{x}_i) \in Y_i )$, and the logarithm of this factor can be seen as the loss caused by $M$ on the observation $(\vec{x}_i , Y_i)$; thus, in GMLI, the counterpart to our generalized loss function (\ref{eq:fuzzyloss}) is given by
\begin{equation}\label{eq:lossd}
\mathbb{L}(Y_i, \hat{y}_i) \, = \, 
- \log \Big( \Prob \left( \widehat{Y}_i \in Y_i \right) \Big) \enspace ,
\end{equation}
where $\widehat{Y}_i$ is a random variable defined by $\hat{y}_i$ and the underlying probabilistic model. More concretely, suppose that $Y_i$ is an interval, say, $Y_i = [3,7]$, and recall our assumption of  an underlying normal distribution (\ref{eq:gaussian}). Then, $\widehat{Y}_i$ is a Gaussian centered at $\hat{y}_i = M(\vec{x}_i)$. As shown in Figure \ref{fig:pic-den-1}, the loss caused by the prediction $\hat{y}_i = 6$ corresponds to the logarithm of the area of this distribution outside the interval $Y_i = [3,7]$.

\begin{figure}
\begin{center}
\includegraphics[scale=0.40]{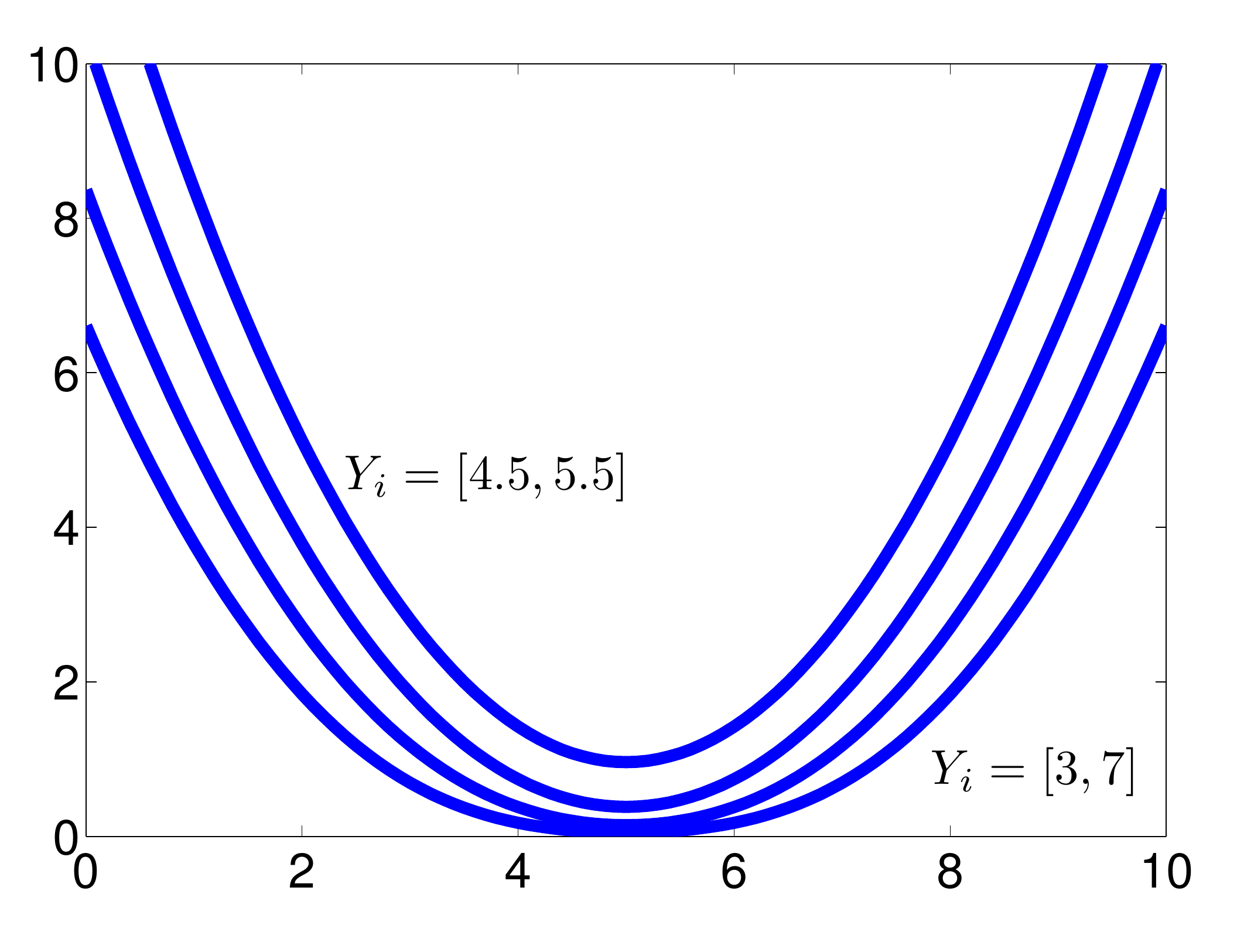} 
\caption{The GMLI loss for intervals of different width around the mid-point $y=5$: $[4.5,5.5]$, $[4,6]$, $[3.5,6.5]$, and $[3,7]$.}
\label{fig:pic-den-2}
\end{center}
\end{figure}

\begin{figure}
\begin{center}
\includegraphics[scale=0.35]{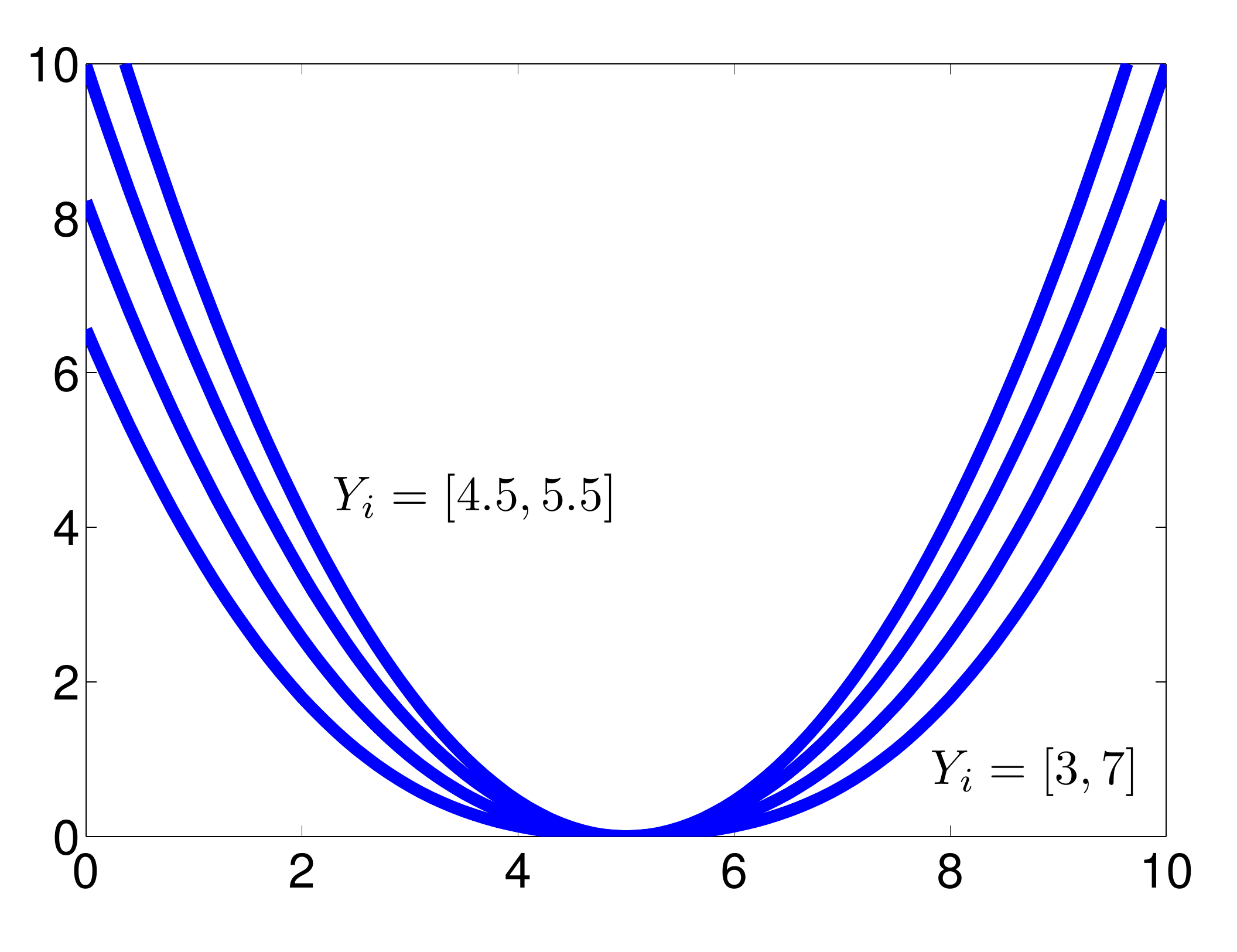} \hspace*{-5mm}
\includegraphics[scale=0.35]{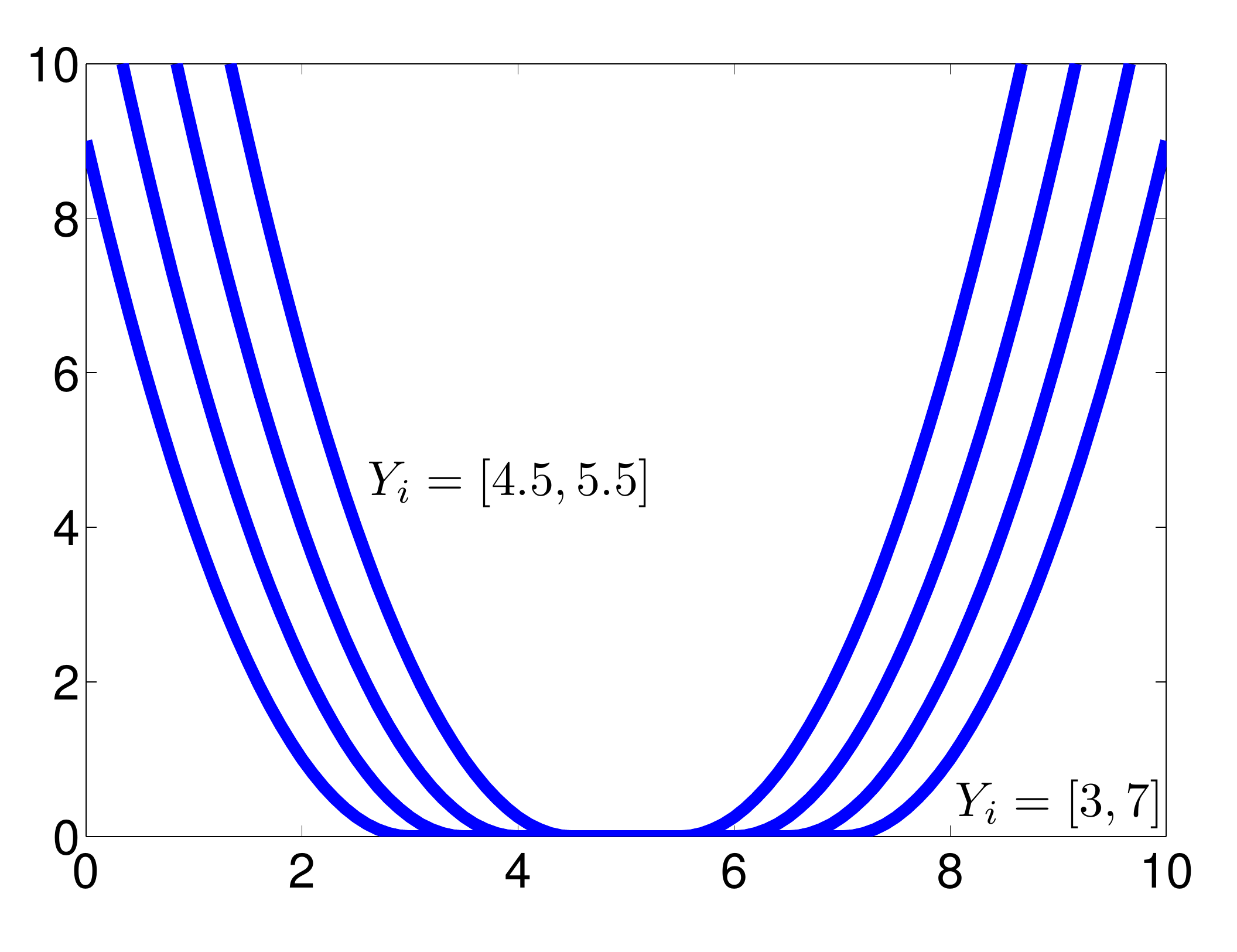}
\caption{Comparison between discounting effects in GMLI and our approach.}
\label{fig:pic-den-3}
\end{center}
\end{figure}

The overall loss function produced in this way (with $\sigma$ in (\ref{eq:gaussian}) given by 1) is shown in Figure \ref{fig:pic-den-2}, together with the loss functions for other intervals (of different width) for comparison. It is noteworthy that the loss in GMLI is never 0, not even when predicting the center of the interval: Even in that case, the Gaussian centered at that value is not completely inside the interval $Y_i$, i.e., $\Prob(M(\vec{x}_i) \in Y_i) < 1$.  

The loss can only become 0 if either $Y_i$ is very large or the Gaussian is very narrow, i.e., if the standard deviation $\sigma$ in (\ref{eq:gaussian}) is very small. This standard deviation, however, is normally estimated globally and not specifically adapted to a single observation; and even if this could be done (the case of heteroscedasticity), the standard deviation would need to be fitted to the data-generating process, not to our \emph{knowledge} about the data. 

Anyway, for a fixed standard deviation, there is a constant and unavoidable penalty that only depends on the width of the interval (and similarly for fuzzy sets): The smaller the interval, the higher the penalty. Please note, however, that this shift of the function does not have any influence on loss minimization: It is simply a constant term in the empirical risk that does not change its minimizer. For better comparison with our approach, we can therefore ``normalize'' the GMLI loss functions by subtracting the constant penalty. 

The result is shown in Figure \ref{fig:pic-den-3}. As can be seen, the discounting of the loss due to an increased imprecision of the observation is quite different in GMLI and our approach: The former is favoring the mid-point of the interval, while the width of the interval (imprecision) leads to a global scaling of the whole loss function: The smaller the interval, the steeper the loss function.\footnote{Note that this may lead to technical problems in the limit case of a precise observation, where the width of the interval tends to 0. Then, even a small prediction error may yield an extreme loss. Obviously, the idea of data inclusion does not naturally apply to the special case of precise data: If $Y_i$ in (\ref{eq:lossd}) reduces from a proper set to a singleton, the probability goes to 0 and hence the logarithm to infinity.} As opposed to this, our approach treats all points inside the interval as equal; likewise, the increase of the loss outside the interval is always the same. Roughly speaking, our approach leads to stretching the loss function ``horizontally'', while GMLI scales it ``vertically''.

\begin{figure}
\begin{center}
\includegraphics[scale=0.35]{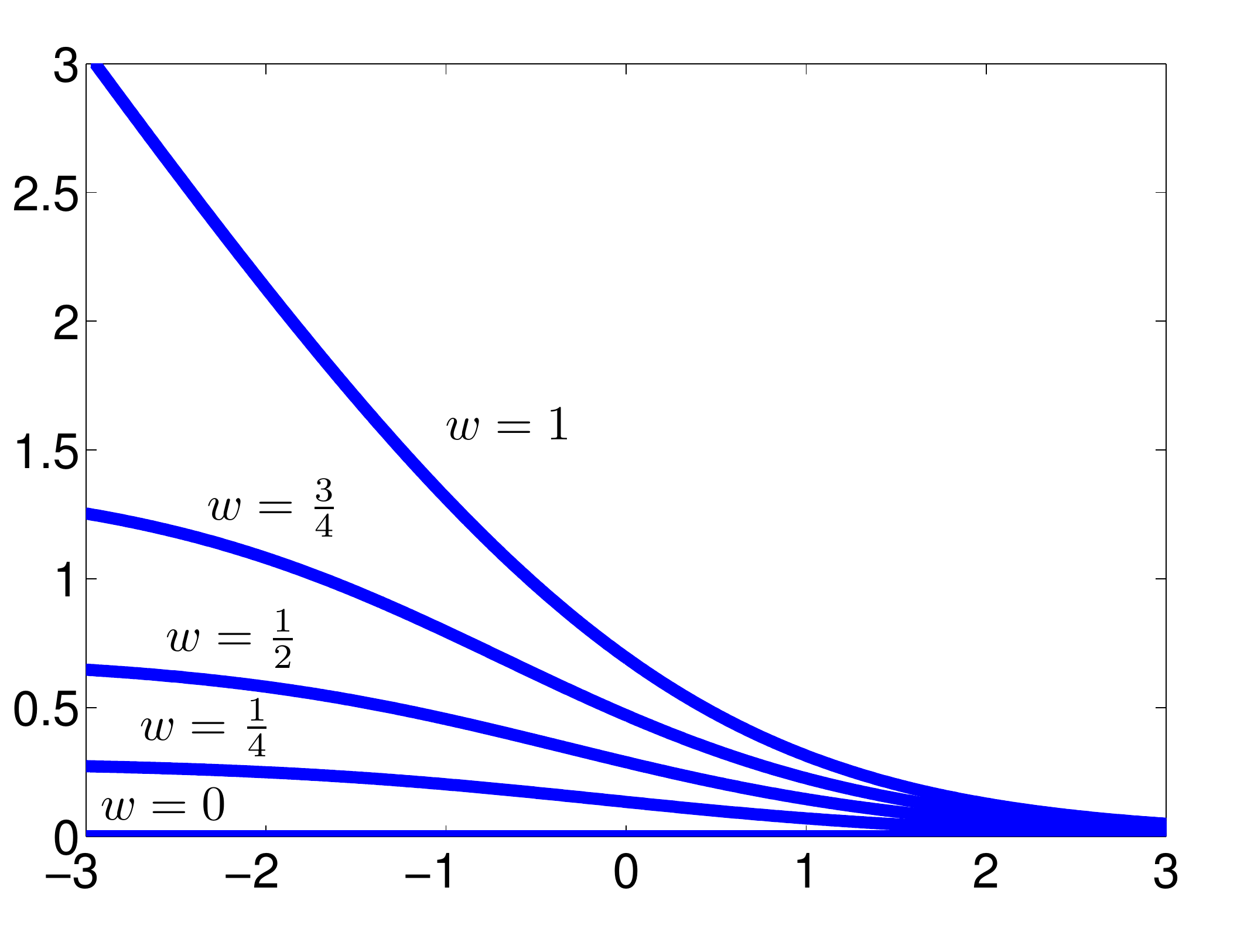} \hspace*{-5mm}
\includegraphics[scale=0.35]{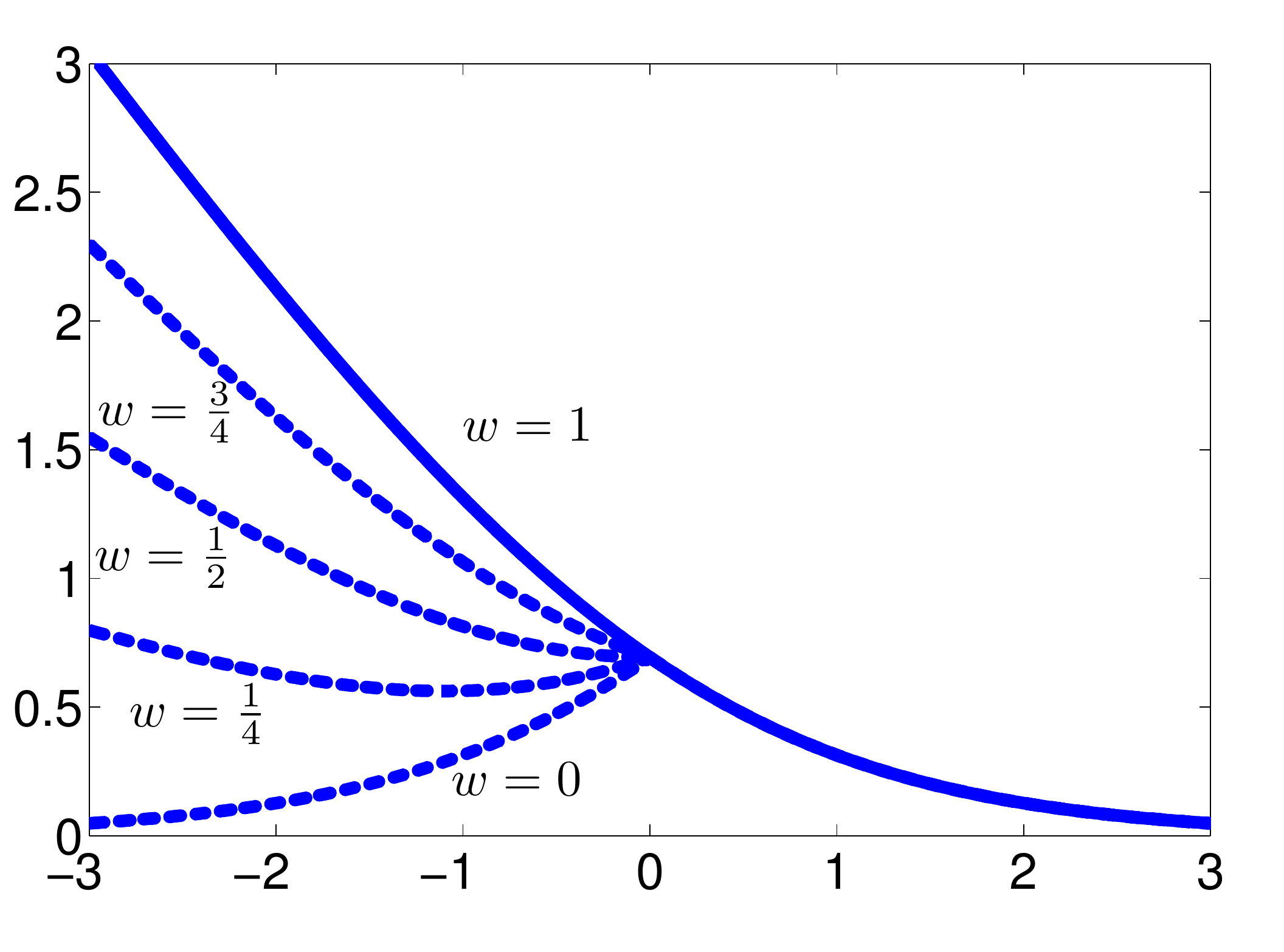}
\caption{Comparison between discounting effects in GMLI and our approach for the logistic loss.}
\label{fig:pic-den-4}
\end{center}
\end{figure}

Qualitative differences of a similar kind can also be seen for other types of loss functions, for example the logistic loss (\ref{logloss}) used in binary classification. For the special type of a discounted (weighted) observation (\ref{eq:disc}), GMLI yields the loss  
\begin{equation}\label{eq:gmlilogloss}
\mathbb{L}(Y, s) = -\log \left( 1 - w \cdot \frac{\exp(-ys)}{1+ \exp(-ys)} \right)
\enspace .
\end{equation}
Figure \ref{fig:pic-den-4} shows these loss functions for different values of $w$ and, for comparison, reproduces the corresponding functions for our approach (already shown in Figure~\ref{fig:mlosses}). In Section~\ref{sec:illu} below, these two loss functions will be compared with each other in a numerical experiment. 

Although the cases we considered here are specific ones, they already suggest that Denoeux's approach is not in agreement with our notion of \emph{data disambiguation}---which is perhaps not surprising, given that it was never intended to implement this idea. In GMLI, the \emph{compatibility} of a model with an imprecise sample is based on the idea of \emph{data inclusion}: When comparing a predicted data point $\hat{\vec{z}}_i$ with an imprecise observation $Z_i$, the loss (log-likelihood) depends on how well $\hat{\vec{z}}_i$ (or, more specifically, the probability distribution associated with that point) is included in $Z_i$. Naturally, this leads to a preference for points ``in the middle'' of $Z_i$. As already mentioned above, the approach therefore tends to fit these middle points, while the imprecision of the information leads to a global decrease of the loss. Our method, on the other hand, starts without any bias in the form of preferences on instantiations and instead tries to figure out the most likely ones.

\section{Illustration}
\label{sec:illu}

This section presents an illustration of our approach in a simple classification setting. Before explaining the setup, we emphasize that our experiments are not meant as an empirical validation of our approach, let alone a comparison with alternative methods in terms of specific performance measures. Since we consider the contribution of this paper as being more of a conceptual than methodological nature, and indeed proposed a conceptual framework rather than a concrete method, such a comparison is arguably not appropriate at this point. 

\begin{figure}
\begin{center}
\includegraphics[scale=0.4]{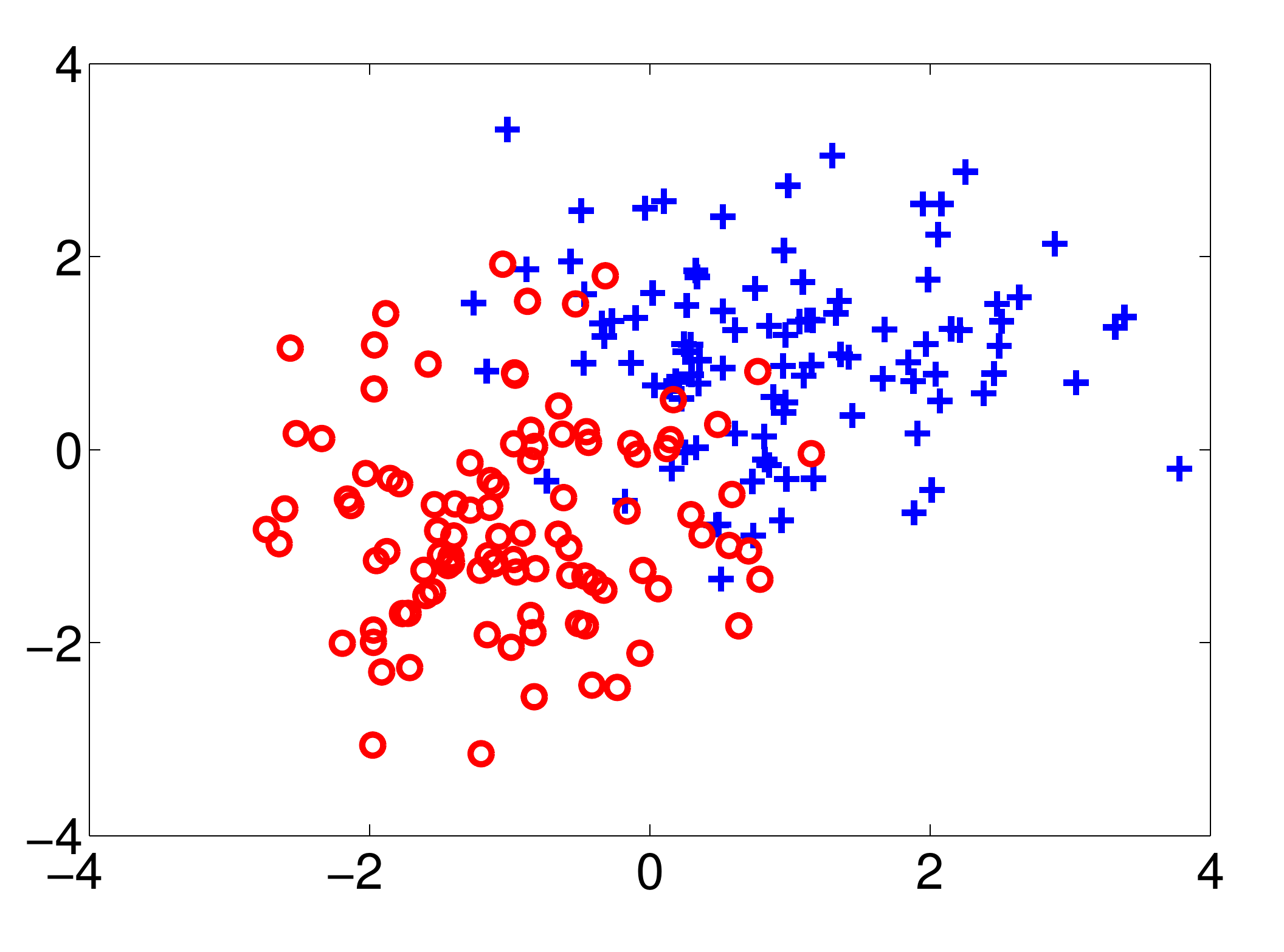} 
\includegraphics[scale=0.4]{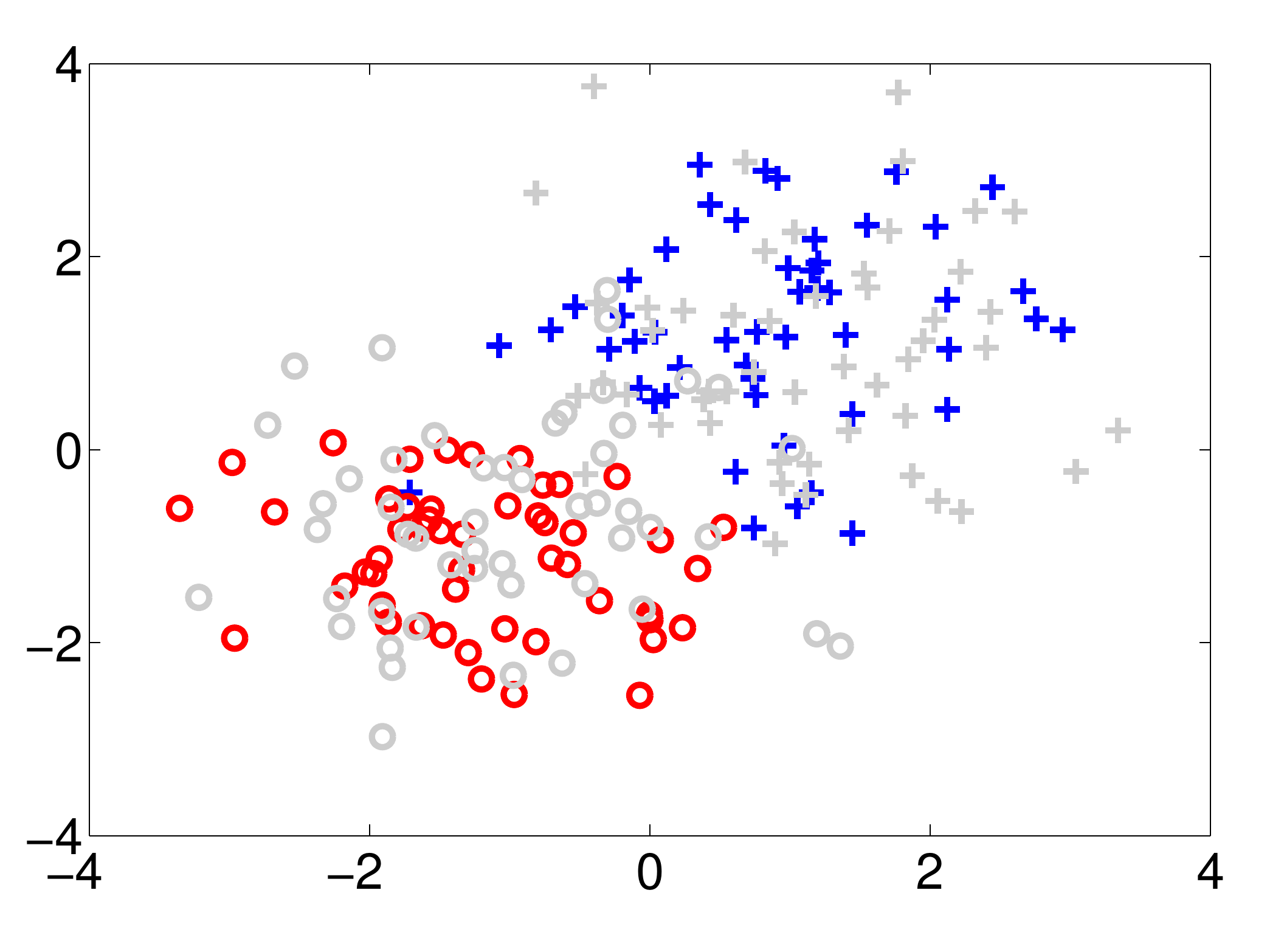}
\includegraphics[scale=0.4]{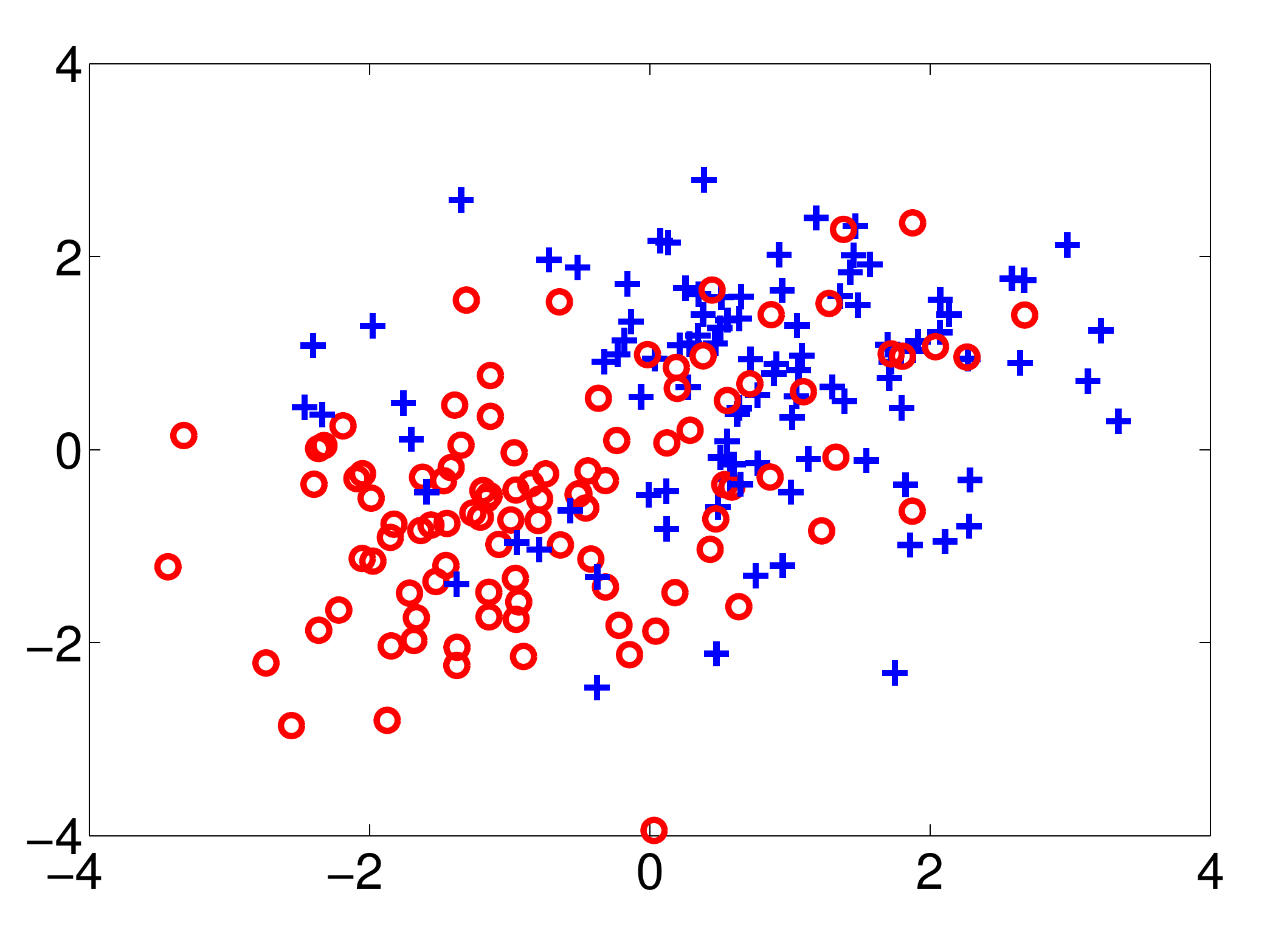}
\caption{(a) Example of a data sample consisting of positive ($+$) and negative ($\circ$) instances in $\mathbb{R}^2$. (b) Example of a data set with ($50\%$) missing class information, indicated in light grey. (c) Example of a data set with ($20\%$) noise.}
\label{fig:example0}
\end{center}
\end{figure}

Nevertheless, we would like to show the potential usefulness of our fuzzy loss functions by means of a practical example. To this end, we consider a simple binary classification problem with normally distributed classes in $\mathbb{R}^2$, the positive one with mean $\mu_+=(1,1)$ and the negative one with mean $\mu_- = (-1,-1)$. As training data, we assume a sample consisting of 100 randomly generated instances from both classes; a typical example is shown in Figure \ref{fig:example0}. On a sample of that kind, we train a linear classifier using logistic regression. Since the true conditional class distributions are known, it is not difficult to determine the generalization performance of such a model in terms of the error rate, i.e., the probability of an incorrect classification (which corresponds to the risk (\ref{eq:risk}) with $L$ the $0/1$ loss). 

In a first experiment, the class information was partly removed from the training instances. More specifically, each of the 200 instances was declared ``unlabeled'' with a fixed probability $\gamma$ (while the original label was kept with probability $1-\gamma$); thus, we are in a setting of \emph{semi-supervised learning}, in which approximately $200(1-\gamma)$ of the instances are labeled (see Figure \ref{fig:example0} for a typical data set of that kind). In our approach, the unlabeled instances can be modeled in terms of a fuzzy set that assigns a membership degree of 1 to both the positive and the negative class. Then, a model is trained using the fuzzy loss function (\ref{fmloss}) with $f$ the log-loss (\ref{logloss}).\footnote{Minimization of the empirical loss was done by means of a simple gradient method, which, due to the non-convexity of the loss, may of course end up in local optima.} Standard logistic regression, on the other hand, cannot directly exploit the unlabeled instances, and therefore only used the remaining labeled ones. 

\begin{figure}
\begin{center}
\includegraphics[scale=0.4]{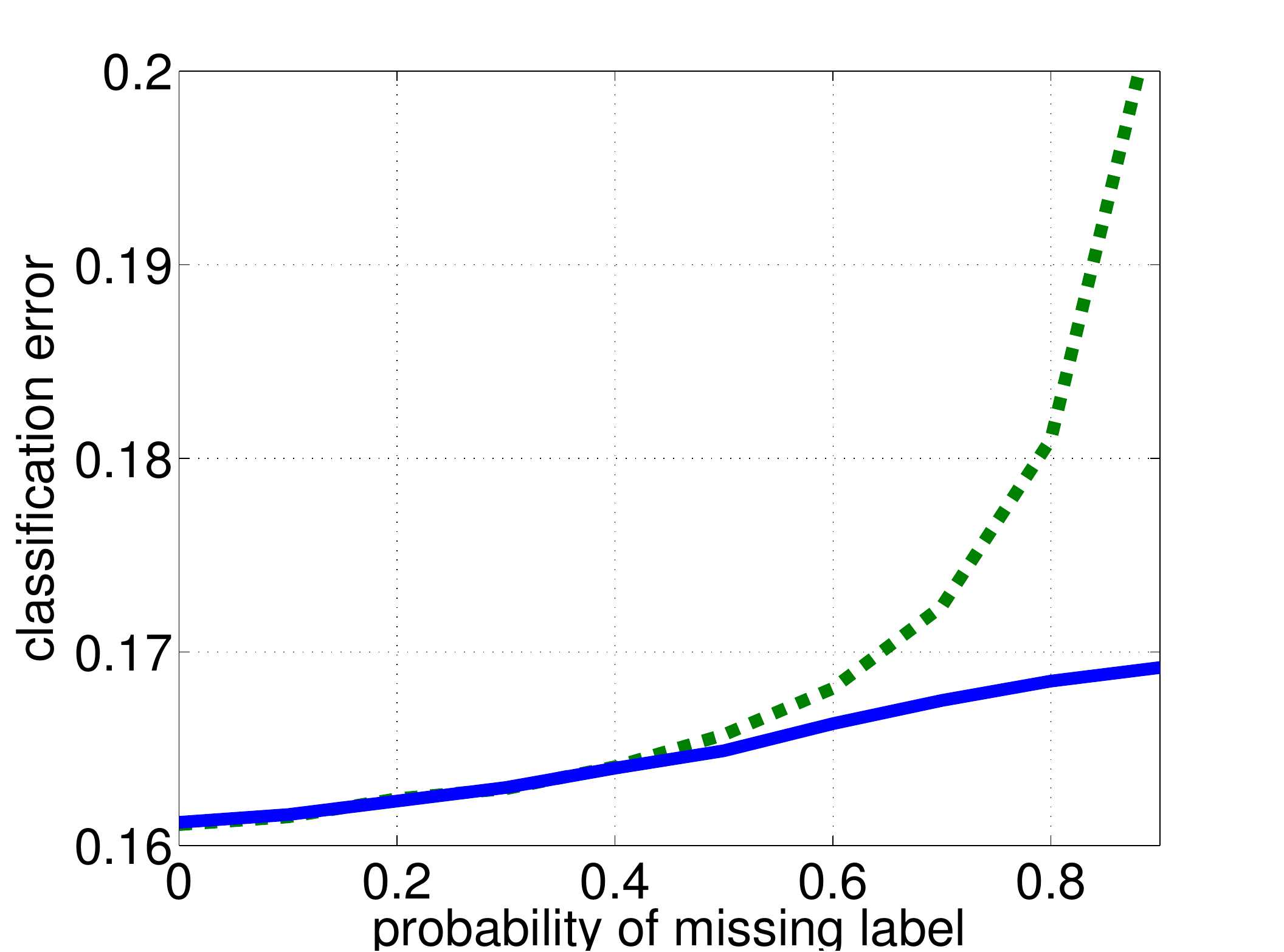} 
\caption{Classification error as a function of the probability of missing label information (first experiment), both for our method (solid line) and standard logistic regression (dashed line).}
\label{fig:example1}
\end{center}
\end{figure}

The results are shown in Figure \ref{fig:example1} in terms of the expected classification error (derived as an average over a large number of repetitions of this experiment) as a function of $\gamma$. As expected, the larger $\gamma$ becomes, i.e., the less labeled and the more unlabeled examples the training data contains, the worse the generalization performance of both methods. Obviously, however, the drop in performance is much more significant for standard logistic regression. From these results, we may conclude that our fuzzy loss (\ref{fmloss}) allows for exploiting the unlabeled instances, in addition to the labeled ones, in a meaningful way. 

As a side remark, we note that Denoeux's GMLI will produce exactly the same result as standard logistic regression: Although the unlabeled data could be modeled in the same way as in our approach, it will effectively be ignored by GMLI: Since the probability to observe either the positive or the negative label (the sure event) is 1, the unlabeled instances will not influence the likelihood function.

\begin{figure}
\begin{center}
\includegraphics[scale=0.4]{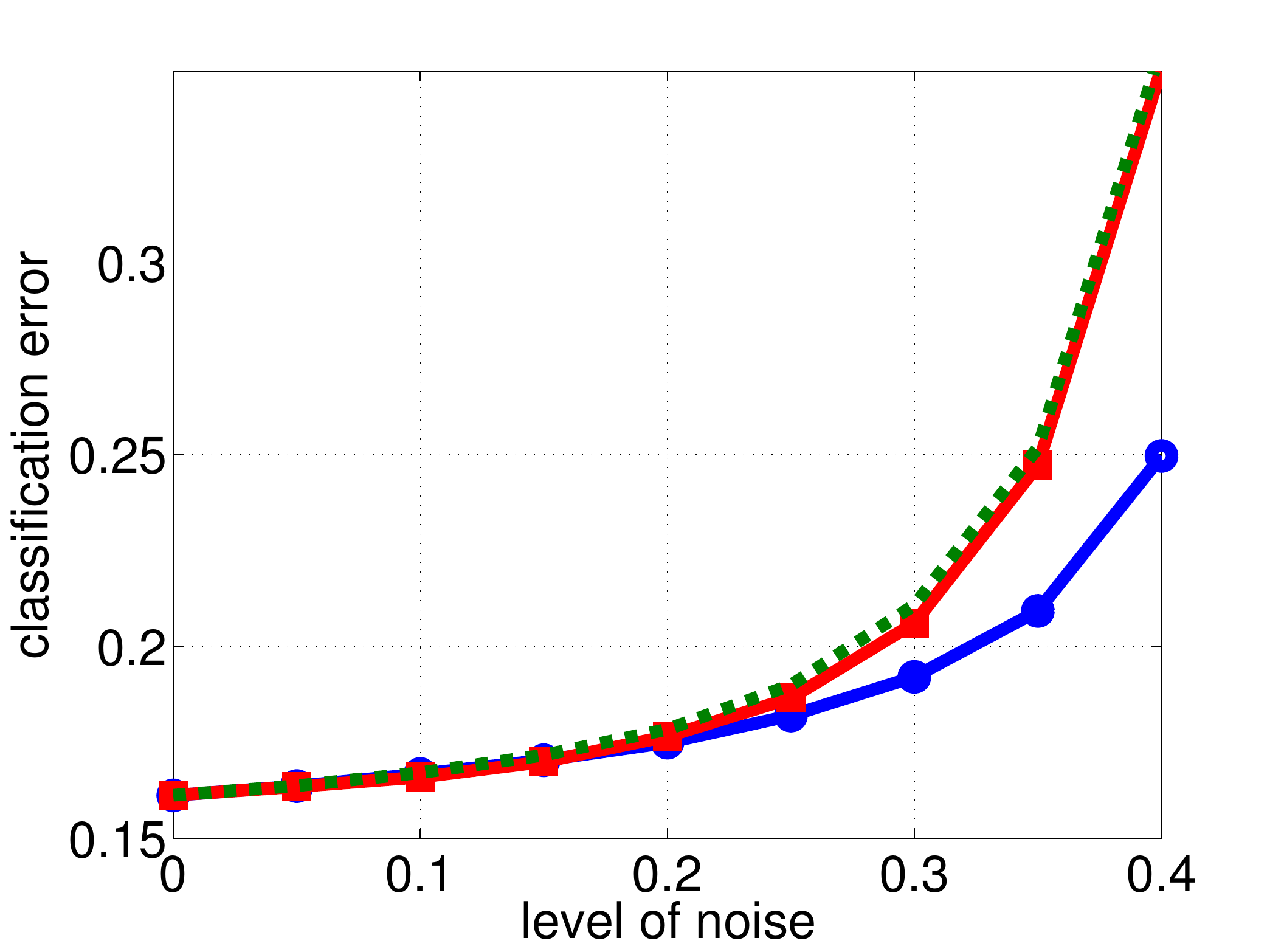} 
\caption{Classification error as a function of the level of noise (second experiment) for standard logistic regression (dashed line), GMLI (solid line, squared markers) and our method (solid line, circle).}
\label{fig:example2}
\end{center}
\end{figure}

In a second experiment, we assume that the label of each example is switched (from positive to negative and vice versa) with a fixed probability $\gamma$, which can be seen as a kind of noise level. This noise level is supposed to be known, whereas for each individual training example, it is not known whether the observed label corresponds to the original one or has been switched. In our approach as well as in GMLI, we can use the idea of attaching a degree of certainty to an observation: The label information is modeled in terms of a fuzzy set (\ref{eq:disc}), assigning a membership degree of 1 to the observed and of $\gamma$ to the other label. For our approach, we again use the fuzzy loss function (\ref{fmloss}) with $f$ the log-loss (\ref{logloss}), whereas GMLI is based on the minimization of the loss (\ref{eq:gmlilogloss}). Standard logistic regression simply uses the observed label information, which is the best it can do. 

Figure \ref{fig:example2} shows the average classification error of the three methods as a function of the noise level $\gamma$. Overall, the picture is quite similar to the first experiment: Compared to our approach, the drop in performance is much more significant for standard logistic regression. This time, GMLI is not exactly equivalent to standard logistic regression, but the difference in performance is negligible. Apparently, our fuzzy loss function (\ref{fmloss}) is more apt to exploit the uncertain training information than the modified loss (\ref{eq:gmlilogloss}) underlying GMLI.

\section{Conclusion}

We have introduced a conceptual framework for (supervised) learning from imprecise and fuzzy data, which is based on the generalization of loss functions in empirical risk minimization. In contrast to the generic extension principle, our approach implicitly exploits the inductive bias underlying the learning method and performs model identification and data disambiguation simultaneously. 

Our extended loss functions allow for directly ``comparing'' a (precise) prediction with an imprecise observation, and thereby provide the basis for fitting a precise model to imprecise data. The principle that we used for extending a standard loss function is coherent with our idea of data disambiguation and can be seen as a sample-specific ``modulation'' of the original loss. 

Interestingly enough, our fuzzy set-based generalization of loss functions covers several existing methods as special cases, including instance weighting, robust regression (Huber loss) and support vector regression ($\epsilon$-insensitive loss). Thus, it may have the potential to serve as a unifying framework of such methods. Apart from that, however, it also allows for deriving new methods in a systematic and conceptually sound manner. For example, while the well-known Huber loss and the $\epsilon$-insensitive loss are obtained by modulating the $L_1$ loss with a symmetric triangular fuzzy set and an interval, respectively, a trapezoidal fuzzy set leads to a new loss function that elegantly combines both effects (insensitivity and robustness) at the same time.

Needless to say, while being conceptually simple, our framework can become quite challenging from a computational perspective. In particular, solving the generalized risk minimization problems (\ref{bestmodel}) and (\ref{bestfuzzymodel}) is far from trivial. Therefore, developing efficient algorithms for specific problem classes is an important topic of future work. Such algorithms will also provide the basis for a proper empirical evaluation of our framework.


\end{document}